\documentclass[10pt,twocolumn,letterpaper]{article}

\usepackage{cvpr}              
\usepackage{multirow}
\usepackage{colortbl}

\definecolor{cvprblue}{rgb}{0.21,0.49,0.74}
\usepackage[pagebackref,breaklinks,colorlinks,allcolors=cvprblue]{hyperref}

\def\paperID{6512} 
\def\confName{CVPR}
\def\confYear{2025}

\title{Point Cloud Understanding via Attention-Driven Contrastive Learning}

\author{%
\textbf{Yi Wang$^{1}$}\thanks{Equal Contribution}
\and
\textbf{Jiaze Wang$^{2*}$} \and
\textbf{Ziyu Guo$^{2}$} \and
\textbf{Renrui Zhang$^{2}$} \and
\textbf{Donghao Zhou$^{2}$} \and
\textbf{Guangyong Chen$^{3}$} \and
\textbf{Anfeng Liu$^{1}$} \and
\textbf{Pheng-Ann Heng$^{2}$}
\\
\\
$^{1}$ Central South Unversity \quad
$^{2}$ The Chinese University of Hong Kong  \quad
${^3}$ Zhejiang Lab
}

\begin{document}
\maketitle
\begin{abstract}
Recently Transformer-based models have advanced point cloud understanding by leveraging self-attention mechanisms, however, these methods often overlook latent information in less prominent regions, leading to increased sensitivity to perturbations and limited global comprehension.
To solve this issue, we introduce \textbf{PointACL}, an attention-driven contrastive learning framework designed to address these limitations. Our method employs an attention-driven dynamic masking strategy that guides the model to focus on under-attended regions, enhancing the understanding of global structures within the point cloud. Then we combine the original pre-training loss with a contrastive learning loss, improving feature discrimination and generalization.
Extensive experiments validate the effectiveness of PointACL, as it achieves state-of-the-art performance across a variety of 3D understanding tasks, including object classification, part segmentation, and few-shot learning. Specifically, when integrated with different Transformer backbones like Point-MAE and PointGPT, PointACL demonstrates improved performance on datasets such as ScanObjectNN, ModelNet40, and ShapeNetPart. This highlights its superior capability in capturing both global and local features, as well as its enhanced robustness against perturbations and incomplete data.
\end{abstract}
\section{Introduction}
Point clouds are widely applicable in fields such as robotics~\citep{chen20203d,tan2001exploring}, autonomous driving~\citep{chen2017multi,chen20203d}, augmented reality~\citep{arena2022overview}, and virtual reality~\citep{garrido2021point} as a representation in three-dimensional space. These diverse applications highlight the significance of obtaining detailed and insightful 3D representations. Despite their potential, the irregular and sparse nature of point cloud data poses significant challenges to precise and efficient 3D processing and understanding.

Recent advancements in deep neural networks, especially Transformer-based models~\citep{pang2022masked,chen2024pointgpt,yu2022point} employing self-supervised learning, have shown promise in point cloud understanding. These models leverage the attention mechanism to capture complex relationships between point patches, prioritizing critical regions for understanding the point cloud while downplaying less significant areas. Originally designed for natural language, attention mechanism has been successfully adapted for 2D vision. However, unlike natural language~\citep{devlin2018bert} or images~\citep{he2022masked}, which often contain redundant information such as contextual structures and backgrounds, point cloud data are inherently sparse, meaning that each point or region is critical to the overall representation. This scarcity of redundant information implies that Transformer-based models, when neglecting less prominent point patches, may inadvertently overlook essential latent information. This observation leads us to a pivotal question: 
\emph{Can we design a framework that leverages latent information from the global regions of point clouds?}

\begin{figure*}[t]
    \centering
        \includegraphics[width=\linewidth]{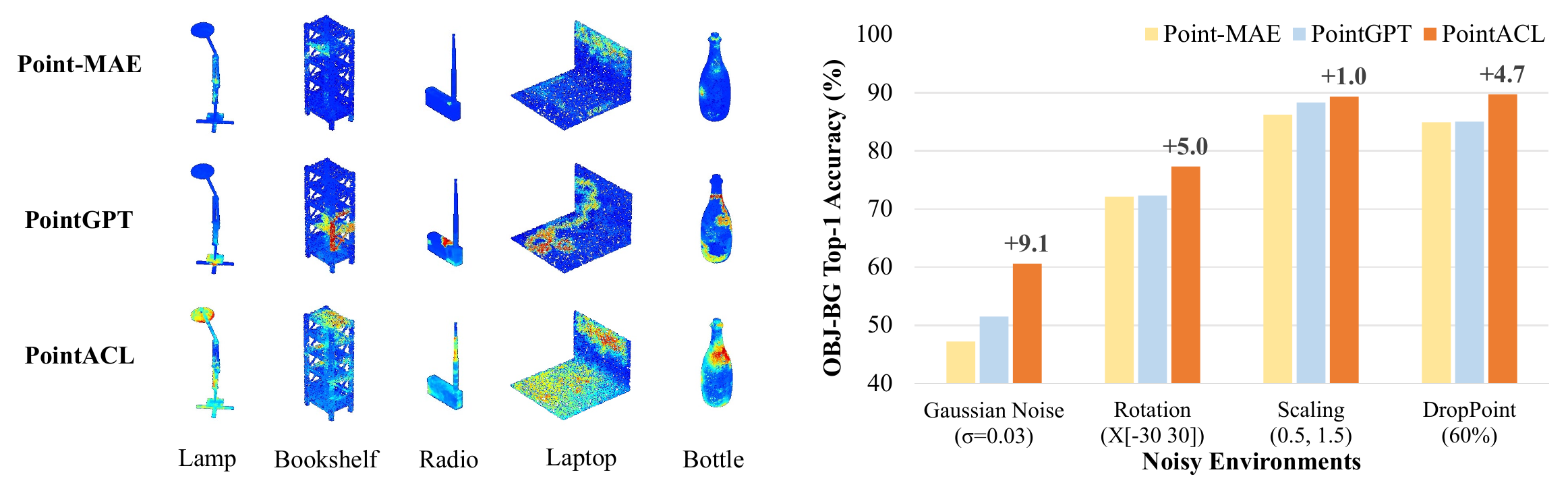}
\caption{\textbf{Illustration of PointACL's Advantages.}  Point-MAE is employed as the backbone of our proposed PointACL. \textbf{Left:} PointACL emphasizes extracting global information from a greater number of patches. \textbf{Right:} PointACL demonstrates greater robustness than previous methods.} 
    \label{fig: performance}
    \vspace{-14pt}
\end{figure*}

To answer this question, we re-examine the attention weights in Transformer-based point cloud models. As illustrated in Figure~\ref{fig: performance}, we find that models like Point-MAE~\citep{yu2022point} and PointGPT~\citep{chen2024pointgpt} primarily rely on a limited set of high-attention patches for analysis. This reliance presents two significant issues: \textbf{(1)} Increased sensitivity to perturbations. Over-focusing on high-attention patches makes the models more susceptible to noise and incomplete data, as disturbances in these areas disproportionately affect performance. \textbf{(2)} Limited global understanding. Ignoring potential information in low-attention patches constrains the model's ability to develop a comprehensive understanding of the point cloud's global structure. 



To solve these issues, we introduce \textbf{PointACL}, an \textbf{A}ttention-driven \textbf{C}ontrastive \textbf{L}earning framework for point clouds that can be seamlessly integrated into existing Transformer-based models. Our approach comprises two key components:
First, an attention-driven dynamic masking strategy is proposed that aims to mitigate the model’s reliance on a limited subset of key patches by guiding it to focus on under-attended regions. Specifically, we construct a dynamic masking probability based on the latest self-attention significance scores, prioritizing masking the patches that contribute most to the global feature representation. This strategy encourages the model to infer global features from less prominent patches, thus fostering a more comprehensive and robust understanding of the point cloud. 
Furthermore, we combine the original pre-training loss with a contrastive learning objective. It allows the model to retain its task-specific learning capabilities while enhancing its global understanding and generalization through contrastive learning. 
Compared to previous methods, our approach better captures the global structure of point clouds rather than focusing solely on local features. Consequently, under various noisy environments such as Gaussian noise, rotation, scaling, and point dropout, PointACL significantly enhances the model's robustness.

 Our PointACL achieves state-of-the-art performance across various 3D understanding tasks. Specifically, for object classification, PointACL attains accuracies of 89.9\% on the challenging PB-T50-RS setting of ScanObjectNN and 94.1\% on ModelNet40, with its performance advantage persisting even when competing models are allocated additional training time. In few-shot learning, it sets new benchmarks across all evaluation tasks. Moreover, PointACL demonstrates enhanced robustness against perturbations and incomplete data, consistently outperforming previous approaches under various noisy environments such as Gaussian noise, rotation, scaling, and point dropout. These results highlight PointACL's potential to effectively address the limitations of existing Transformer-based models by capturing comprehensive global structures and fine-grained local details.

Our main contributions can be summarized as follows: 

\textbf{(I)} We propose PointACL, a novel framework that combines self-attention mechanisms with contrastive learning for point cloud understanding which enhances the model's ability to capture global structures and significantly improves its robustness and generalization capabilities.

\textbf{(II)} We propose an attention-driven dynamic masking strategy that encourages the model to focus on under-attended regions, ensuring learning from diverse patches rather than over-relying on a small subset.

\textbf{(III)} Extensive experimental results demonstrate that PointACL can be seamlessly integrated into mainstream transformer architectures and achieve significant improvements across a variety of 3D understanding tasks.
\section{Related Works}

\noindent\textbf{Self-Supervised Learning for NLP and Image.}
Self-supervised learning (SSL) has emerged as a powerful paradigm in natural language processing (NLP)~\citep{erhan2010does,zhu20233d} and computer vision~\citep{radford2018improving,goodfellow2020generative,yu2017seqgan,misra2020self,qian2021spatiotemporal,abdelfattah2024maskclr, liang2024pointmamba}, enabling models to learn rich representations from unlabeled data. 
In NLP, BERT~\citep{devlin2018bert} exemplifies this by randomly masking input tokens and training the model to predict them, fostering deep contextual understanding. ELMo~\citep{sarzynska2021detecting} utilizes bidirectional LSTMs to generate contextualized word embeddings, while GPT~\citep{radford2018improving} adopts an autoregressive approach with a unidirectional Transformer to predict the next word, fine-tuning all parameters for specific tasks.
Recently, generative SSL methods have begun to outperform contrastive approaches in computer vision. Masked Autoencoders~\citep{he2022masked} randomly mask image patches and train the model to reconstruct the missing pixels, leading to effective visual representations. BEiT~\citep{bao2021beit} extends this by tokenizing image patches and predicting masked tokens, integrating NLP techniques into vision tasks. Additionally, Image GPT~\citep{luppino2021deep} treats images as sequences of pixels and trains a Transformer to autoregressively predict pixels without explicit spatial structure, demonstrating strong representation learning. This shift towards generative self-supervised learning methods not only demonstrates their ability to capture comprehensive data representations and improve performance in NLP and computer vision but also highlights their significant potential in advancing point cloud processing and analysis.

\noindent\textbf{Self-Supervised Learning for Point Cloud.} Various methods have been investigated for self-supervised learning on point clouds ~\citep{wang2024pointramba, liu2024point,wu2024point,zhang2023learning, zhang2024point, han2024mamba3d, zha2024towards, feng2024shape2scene}.
Many works focused on generative modeling with generative adversarial networks and autoencoders, aiming to reconstruct inputs using different architectural designs~\citep{min2022voxel,yu2022point,sauder2019self,li2018so,achlioptas2018learning,wang2022p2p}.
PointMAE~\citep{pang2022masked} proposes an effective scheme of masked autoencoders for point cloud self-supervised learning.
Point-M2AE~\citep{zhang2022point} further employs a hierarchical transformer architecture and implements a specific masking strategy.
PointGPT~\citep{chen2024pointgpt} propose a point cloud auto-regressive generation task to pre-train transformer models.
Moreover, contrastive methods also have been extensively explored~\citep{qian2022pointnext,xue2023ulip,xue2024ulip,navaneet2020image,zhang2021self,xie2020pointcontrast,huang2023clip2point}.
DepthContrast~\citep{zhang2021self} generates augmented depth maps and conducts instance discrimination on the extracted global features. MVIF~\citep{jing2020self} employs cross-modal and cross-view invariance constraints to enable self-supervised learning of modal- and view-invariant features.
OcCo~\citep{wang2021unsupervised} aims to reconstruct the original point cloud from an occluded version observed in camera views.
Some studies focus on integrating cross-modal information, utilizing knowledge from language or image models to enhance 3D learning~\citep{qi2023contrast,dong2022autoencoders, saito2024point, qi2025shapellm}.
PointCLIP~\citep{zhang2022pointclip} facilitates the alignment between point clouds encoded by CLIP and corresponding 3D category text descriptions, enhancing cross-modal understanding.
PointCLIP V2~\citep{zhu2023pointclip} uses a shape projection module to guide CLIP in generating more realistic depth maps and prompts a GPT model to create 3D-specific text for CLIP's textual encoder input.
Our work bridges the gap between prior self-supervised learning methods and advanced point cloud processing techniques, contributing to the progression of 3D point cloud analysis.
\begin{figure*}[t]
    \centering
        \includegraphics[width=\linewidth]{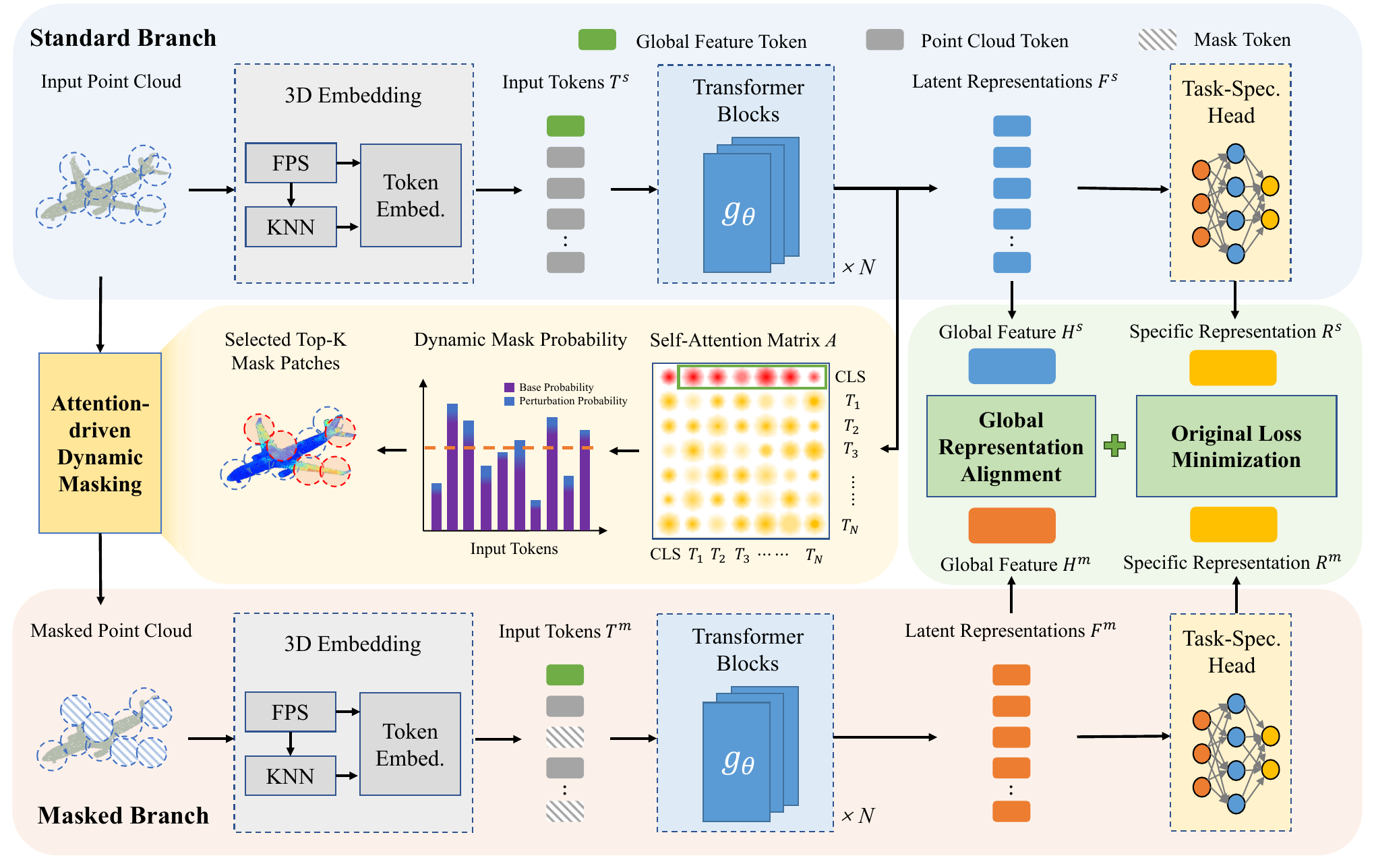}
\caption{\textbf{Overview of the PointACL Framework.} PointACL consists of two branches that share the same weights: a standard mode branch and a masked mode branch. An attention-driven dynamic masking module generates a masked point cloud by selecting less activated patches from the output of the standard mode branch. Both branches process their respective inputs through the shared Transformer blocks to obtain latent representations. Finally, a joint contrastive loss is used to align the representations of these two branches.}
    \label{fig: main}
    \vspace{-10pt}
\end{figure*}

\section{Methods}
The overall framework of PointACL is illustrated in Figure \ref{fig: main}. First, the Attention-driven Dynamic Masking module generates an attention-guided masked point cloud. Both the masked point cloud and the original input point cloud are then fed into the shared backbone model to obtain the global features of each input. By aligning the features from these two branches with contrastive loss, we guide the model to focus on the low-attention regions of the point clouds, thereby improving feature discrimination and generalization. During the pre-training stage, we train the model using a combination of contrastive loss and the original pre-training loss—such as the reconstruction loss from PointMAE~\citep{pang2022masked} or the generation loss from PointGPT~\citep{chen2024pointgpt}. After pre-training, we employ the backbone model without the masking strategy, leveraging the learned latent representations for downstream tasks.

\subsection{Attention-Driven Dynamic Masking}

To fully harness the advantages of the self-attention mechanism and mitigate the model's reliance on a small subset of key patches, we propose an attention-driven dynamic masking strategy, which guides the model to focus on low-attention regions and enforces a more comprehensive understanding of the global structure in challenging scenarios by dynamically masking high-attention areas.

\noindent\textbf{Point patch attention.}
Given a point cloud $X\in \mathbb{R}^{P\times 3}$, we utilize Farthest Point Sampling (FPS) and K-Nearest Neighbors (KNN) algorithms to identify $n$ center points $C$ and their corresponding $k$ nearest neighbors, forming $n$ point patches $P$.
We employ the self-attention mechanism in the transformer architecture to compute the attention weights of point patches relative to the global feature. A new set of input tokens $T\in\mathbb{R}^{(N+1)\times d}$, consisting of the point tokens $T^p\in\mathbb{R}^{N\times d}$ and a learnable global feature token $T^f\in\mathbb{R}^{1\times d}$, is utilized to compute the queries $Q\in\mathbb{R}^{(N+1)\times d}$ , keys $K\in\mathbb{R}^{(N+1)\times d}$ , and values $V\in\mathbb{R}^{(N+1)\times d}$. The attention matrix $A$ is subsequently derived from the dot product of the queries and keys.
Since the first element of the input tokens $T_1$ corresponds to the global feature token, the first row of the attention matrix can be interpreted as the contribution of each token to the global feature. Considering the output tokens depend on both the attention matrix and the values, we incorporate the norm of $V_j$ when determining the significance score of token $j$. Consequently, the attention matrix and significance score for point patch $j$ are computed as follows:
\begin{equation}
A = Softmax(QK^{T}/\sqrt{d}),
\end{equation}
\begin{equation}
S_{j}=\frac{A_{1,j}\times \left \| V_{j} \right \| }{ {\textstyle \sum_{i=2}A_{1,i}\times\left \| V_{i} \right \|  } },
\end{equation}
 where $i,j\in 2, ..., N+1$. For a multi-head attention layer, we compute the significance scores for each head separately and aggregate them by taking the sum over all heads.

\noindent\textbf{Dynamic Masking.} A straightforward idea is to mask the top $k$ patches with the highest significance scores, as they are key to the model's understanding of the point cloud. However, a fixed masking probability merely shifts the model’s attention without engaging a broader set of patches. As the model becomes reliant on new areas of focus, it similarly falls into the trap of limited comprehension of the point cloud. Our primary objective is to ensure that high-attention regions have a higher likelihood of being masked. Therefore, we propose a dynamic masking strategy. Specifically, we construct an updatable base masking probability using the latest self-attention significance scores, prioritizing the masking of patches that currently contribute significantly to the global features. Additionally, a perturbation probability, derived from a uniform distribution $U[0,1]$, is introduced to enhance the variability of the masking probability. Based on this concept,  the final dynamic masking probability $p_{dy}$ is expressed as:
\begin{equation}
p_{dy} = \log\left(Softmax(S/\tau_{pro})\right)-\log\left(-\log\varepsilon\right),
\end{equation}
where $\tau_{pro}$ is a temperature hyperparameter which controls the sharpness of the base masking probability. A lower temperature (less than 1) results in a sharper distribution, meaning that regions with the highest attention are more likely to be masked. Based on the dynamic masking probability, we apply simple Top-K strategy to select the $k$ point patches $P^{mask}\in \mathbb{R}^{K\times 3}$ to be masked:
\begin{equation}
P^{mask} = \text{Top-K}(p_{dy}, k).
\end{equation}
They are then replaced with learnable mask tokens. In this manner, regions that attract high attention are more likely to be masked, promoting a deeper understanding of the global structure by the model.

\subsection{Learning Objective}
To further improve the model's feature discrimination and generalization, we introduce the contrastive loss to the pre-training stage, which combines the original pre-training loss with a contrastive learning objective, enabling the model to retain task-specific learning capabilities while enhancing its global understanding.

\noindent\textbf{Global representation alignment.} The dynamically selected masked token $T^{m}$ and the standard token $T^{s}$ are both input into a shared-weight model, producing two distinct levels of point cloud latent representations $F^{m}$ and $F^{s}$. Unlike the masked latent representations, the complete point cloud retains all original information. Although the masking strategy results in the loss of some regional details, both representations still correspond to the same underlying point cloud entity. Therefore, we expect the global features extracted from the masked point cloud to align with those derived from the standard point cloud. This alignment ensures that the model becomes less dependent on specific, localized regions and is encouraged to focus on the global structural characteristics of the point cloud. To achieve this, we introduce a contrastive learning objective:
\begin{equation}
\begin{split}
\mathcal{L}_{contra} = -\frac{1}{2b} \sum_{i}
&(
\log\frac{\exp(H^{m}_i \cdot H^{s}_i / \tau_{sim})}{\sum_{j} \exp(H^{m}_i \cdot H^{s}_j / \tau_{sim})} + \\
&\log\frac{\exp(H^{s}_i \cdot H^{m}_i / \tau_{sim})}{\sum_{j} \exp(H^{s}_i \cdot H^{m}_j / \tau_{sim})}),
\end{split}
\end{equation}
where $b$ is the number of point clouds in a batch; $\tau_{sim}$ is a temperature hyperparameter; $H^{m}_i$ and $H^{s}_i$ are the normalized projection features of $F^{m}_i$ and $F^{s}_i$. By omitting the high-attention regions in the masked point clouds, the contrastive objective incentivizes the model to focus on and extract valuable information from less emphasized areas. This process facilitates the learning of a more holistic latent representation, thereby improving the model’s capacity to effectively differentiate between various point cloud objects.

\noindent\textbf{Contrastive learning enhancement.} While traditional contrastive learning methods have demonstrated significant success in unsupervised and self-supervised learning, relying solely on contrastive loss may weaken the model's performance on specific tasks. This limitation arises from the model’s inability to fully exploit the advantages of the existing framework. To address this issue, we propose that a better solution is to integrate the contrastive loss into the existing framework. This approach preserves the model’s task-specific learning capabilities while leveraging contrastive learning to further improve its global understanding and generalization capacity. The proposed total loss is formulated as follows:
\begin{equation}
\mathcal{L}_{total} = \mathcal{L}_{origin} + \lambda\mathcal{L}_{contra},
\end{equation}
where $\mathcal{L}_{origin}$ represents the original loss in the existing framework; $\lambda$ is a weight hyperparameter that controls the contribution of contrastive learning loss. During the pre-training phase, Point-MAE's original pre-training loss $\mathcal{L}_{origin}$ is equivalent to the reconstruction loss $\mathcal{L}_{re}$. For PointGPT, $\mathcal{L}_{origin}$ refers to the generation loss $\mathcal{L}_{ge}$. Therefore, we jointly optimizes the reconstruction (or generation) and contrastive losses, ensuring that the model not only achieves high-quality reconstructions (or generations) but also learns globally consistent feature representations. Through this strategy, PointACL exhibits strong potential for adaptability and scalability across a wide range of multi-task learning scenarios, ultimately improving the model's overall performance.

\section{Experiments}
\subsection{Experimental Setup}

\noindent\textbf{Datasets.}
We evaluate PointACL framework on three benchmark datasets commonly used in 3D point cloud analysis. 
\emph{\textbf{ScanObjectNN}}~\citep{uy2019revisiting} comprises approximately 15,000 real-world 3D objects from 15 categories derived from indoor RGB-D scans, presenting challenges like background clutter, occlusions, and sensor noise.
\emph{\textbf{ModelNet40}}~\citep{wu20153d} is a synthetic dataset with 12,311 CAD models across 40 categories, split into 9,843 for training and 2,468 for testing.
\emph{\textbf{ShapeNetPart}}~\citep{yi2016scalable} contains 16,881 models across 16 categories, each annotated with part labels totaling 50 classes, enabling evaluation of fine-grained part segmentation.

\noindent\textbf{Backbone models.}
To evaluate the seamless integration of the proposed method into existing Transformer-based models for point cloud processing, we employed different backbone architectures, specifically Point-BERT, Point-MAE and PointGPT-S, to validate the algorithm's effectiveness. Experimental results across various tasks indicate that the method is adaptable and enhances the performance of these Transformer architectures, thereby demonstrating its versatility and practical applicability.

\noindent\textbf{Experimental details.}
Our input point clouds are obtained by sampling 1,024 points from each raw point cloud. Each point cloud is then divided into 64 patches with 32 points each. The PointACL model is pre-trained for a total of 600 epochs: the first 300 epochs focus on the original task alone, and the next 300 epochs incorporate both original pre-training and contrastive learning objectives. We use the Adam optimizer with an initial learning rate of 0.001, a weight decay of 0.05, and a batch size of 128. The learning rate is adjusted using a cosine decay schedule. All experiments are implemented using the PyTorch framework and conducted on four NVIDIA V100 GPUs. More training strategy details and training cost analysis are provided in the supplementary material.

\begin{table*}[t]
\centering
\small
\caption{\textbf{Object classification on ScanObjectNN and ModelNet40.} We report the Top-1 classification accuracy (\%) of PointACL with Point-MAE and PointGPT-S as backbones respectively. On ScanObjectNN, * denotes using simple rotational augmentation for training. On ModelNet40, * denotes the results obtained by voting.}
\label{table:cls}
\begin{tabular*}{\linewidth}{@{\hspace{0.5em}}@{\extracolsep{\fill}}lccccc}
\toprule
\multirow{2}{*}{Methods} & \multirow{2}{*}{Reference} & \multicolumn{3}{c}{ScanObjectNN} & \multirow{2}{*}{ModelNet40} \\ \cmidrule{3-5}
 &  & OBJ-BG & OBJ-ONLY & PB-T50-RS &  \\ \cmidrule{1-6}
\multicolumn{6}{c}{\emph{Supervised Learning Only}} \\ \cmidrule{1-6}
PointNet~\citep{qi2017pointnet} & CVPR 17 & 73.3 & 79.2 & 68.0 & 89.0 \\
PointNet++~\citep{qi2017pointnet++} & NeurIPS 17 & 82.3 & 84.3 & 77.9 & 90.2 \\
PointCNN~\citep{li2018pointcnn} & NeurIPS 18 & 86.1 & 85.5 & 78.5 & 91.7 \\
DGCNN~\citep{wang2019dynamic} & TOG 19 & 82.8 & 86.2 & 78.1 & 92.0 \\
PRANet~\citep{cheng2021net} & TIP 21 & - & - & 81.0 & 92.9 \\
MVTN~\citep{hamdi2021mvtn} & ICCV 21 & - & - & 82.8 & 93.8 \\
PointNeXt~\citep{qian2022pointnext} & NeurIPS 22 & - & - & 87.7 & 92.9 \\
PointMLP~\citep{ma2022rethinking} & ICLR 22 & - & - & 85.4 & 94.1 \\
RepSurf-U~\citep{ran2022surface} & CVPR 22 & - & - & 84.3 & 93.8 \\
ADS~\citep{hong2023attention} & ICCV 23 & - & - & 87.5 & 94.0 \\ \cmidrule{1-6}
\multicolumn{6}{c}{\emph{with Self-Supervised Representation Learning}} \\ \cmidrule{1-6}
MaskPoint~\citep{liu2022masked} & CVPR 22 & 89.3 & 88.1 & 84.3 & 92.6 \\
Point-M2AE~\citep{zhang2022point} & NeurIPS 22 & 91.2 & 88.8 & 86.4 & 93.4 \\
PointDif~\citep{zheng2024point} & CVPR 24 & 93.3 & 91.9 & 87.6 & - \\ 
GPM~\citep{li2024general} & CVPR 24 & 90.2 & 90.0 & 84.8 & 93.3 \\
PointMamba~\citep{liang2024pointmamba} & NeurIPS 24 &  90.7 & 88.5 & 84.9 & 93.6 \\
\cmidrule{1-6}
Point-BERT~\citep{yu2022point} & CVPR 22 & 87.4 & 88.1 & 83.1 & 92.7 \\
\textbf{+PointACL} & - & \hspace{0.85cm}89.5 \textcolor{blue}{(+2.1)} & \hspace{0.85cm}88.6 \textcolor{blue}{(+0.5)} & \hspace{0.85cm}84.5 \textcolor{blue}{(+1.4)} & \hspace{0.85cm}93.1 \textcolor{blue}{(+0.4)} \\
\cmidrule{1-6}
Point-MAE~\citep{pang2022masked} & ECCV 22 & 90.0 & 88.3 & 85.2 & 93.2 \\
\textbf{+PointACL} & - & \hspace{0.85cm}90.9 \textcolor{blue}{(+0.9)} & \hspace{0.85cm}88.8 \textcolor{blue}{(+0.5)} & \hspace{0.85cm}85.4 \textcolor{blue}{(+0.2)} & \hspace{0.85cm}93.7 \textcolor{blue}{(+0.5)} \\
\cmidrule{1-6}
PointGPT-S~\citep{chen2024pointgpt} & NeurIPS 23 & 91.6 & 90.0 & 86.9 & 93.3 \\
\textbf{+PointACL} & - & \hspace{0.85cm}92.3 \textcolor{blue}{(+0.7)} & \hspace{0.85cm}91.6 \textcolor{blue}{(+1.6)} & \hspace{0.85cm}87.1 \textcolor{blue}{(+0.2)} & \hspace{0.85cm}93.5 \textcolor{blue}{(+0.2)}\\
\cmidrule{1-6}
Point-MAE*~\citep{pang2022masked} & ECCV 22 & 92.8 & 91.2 & 89.0 & 93.8 \\
\textbf{+PointACL}* & - & \hspace{0.85cm}93.1 \textcolor{blue}{(+0.3)} & \hspace{0.85cm}91.7 \textcolor{blue}{(+0.5)} & \hspace{0.85cm}89.2 \textcolor{blue}{(+0.2)} & \hspace{0.85cm}\textbf{94.1} \textcolor{blue}{(+0.3)}\\
\cmidrule{1-6}
PointGPT-S*~\citep{chen2024pointgpt} & NeurIPS 23 & 93.4 & 92.4 & 89.2 & 94.0 \\
\textbf{+PointACL}* & - & \hspace{0.80cm} \textbf{94.5} \textcolor{blue}{(+1.1)} & \hspace{0.80cm} \textbf{93.5} \textcolor{blue}{(+1.1)}& \hspace{0.80cm} \textbf{89.9} \textcolor{blue}{(+0.7)} & \hspace{0.80cm} \textbf{94.1} \textcolor{blue}{(+0.1)}\\
\bottomrule
\end{tabular*}
\vspace{-10pt}
\end{table*}

\subsection{Experimental Results}

\noindent\textbf{Real-world object classification on ScanObjectNN.}
Table~\ref{table:cls} compares our proposed PointACL method with existing approaches on the ScanObjectNN dataset across OBJ-BG, OBJ-ONLY, and PB-T50-RS settings. 
Our PointACL consistently outperforms these state-of-the-art methods. Compared to Point-MAE~\citep{pang2022masked}, PointACL achieves higher accuracies by +0.9\%, +0.5\%, and +0.2\% on OBJ-BG, OBJ-ONLY, and PB-T50-RS, respectively. Against PointGPT-S~\citep{chen2024pointgpt}, PointACL attains improvements of +0.7\%, +1.6\%, and +0.2\% on the same splits. With simple rotational augmentation (marked with *), PointACL sets new state-of-the-art results, achieving up to 94.5\% on OBJ-BG, 93.5\% on OBJ-ONLY and 89.9\% on PB-T50-RS.
These results demonstrate that PointACL effectively enhances feature representation for point cloud data, particularly in challenging scenarios with background noise and object perturbations. The consistent performance gains across all settings highlight the robustness and efficacy of our approach.

\noindent\textbf{Synthetic object classification on ModelNet40.}
Table~\ref{table:cls} presents the performance of our proposed PointACL method compared to existing self-supervised learning approaches on the ModelNet40 dataset, evaluated both without voting and with voting. Our PointACL achieves an accuracy of 93.7\% without voting and 94.1\% with voting, surpassing previous methods without adding additional parameters. Specifically, compared to Point-MAE, PointACL improves accuracy by +0.5\% without voting and +0.3\% with voting. When compared to PointGPT-S, our method achieves gains of +0.2\% and +0.1\%, respectively. These results demonstrate that PointACL effectively enhances feature representation learning for 3D point cloud data, leading to superior classification performance on ModelNet40.

\begin{table}[t]
\centering
\caption{\textbf{Few-shot classification on ModelNet40.} We report the mean accuracy (\%) with standard deviation over 10 independent experiments.}
\label{table:few}
\resizebox{\linewidth}{!}{
\begin{tabular}{lcccc}
\toprule
\multirow{2}{*}{Methods} & \multicolumn{2}{c}{5-way} & \multicolumn{2}{c}{10-way} \\ \cmidrule{2-5}
& 10-shot & 20-shot & 10-shot & 20-shot \\ \cmidrule{1-5}
\multicolumn{5}{c}{\emph{Supervised Learning Only}} \\ \cmidrule{1-5}
PointNet & 52.0±3.8 & 57.8±4.9 & 46.6±4.3 & 35.2±4.8 \\
PointNet-CrossPoint & 90.9±1.9 & 93.5±4.4 & 84.6±4.7 & 90.2±2.2 \\
DGCNN & 31.6±2.8 & 40.8±4.6 & 19.9±2.1 & 16.9±1.5 \\
DGCNN-CrossPoint & 92.5±3.0 & 94.9±2.1 & 83.6±5.3 & 87.9±4.2 \\ \cmidrule{1-5}
\multicolumn{5}{c}{\emph{with Self-Supervised Representation Learning}} \\ \cmidrule{1-5}
MaskPoint & 95.0±3.7 & 97.2±1.7 & 91.4±4.0 & 93.4±3.5 \\
Point-M2AE & 96.8±1.8 & 98.3±1.4 & 92.3±4.5 & 95.0±3.0 \\ \cmidrule{1-5}
Point-BERT & 94.6±3.1 & 96.3±2.7 & 91.0±5.4 & 92.7±5.1 \\
\textbf{+PointACL} & \textbf{95.1±2.6} & \textbf{97.4±2.0} & \textbf{91.5±4.7} & \textbf{93.0±3.1} \\ \cmidrule{1-5}
Point-MAE & 96.3±2.5 & 97.8±1.8 & 92.6±4.1 & 95.0±3.0 \\
\textbf{+PointACL} & \textbf{96.7±2.7} & \textbf{98.2±1.6} & \textbf{92.8±4.0} & \textbf{95.3±3.2} \\ \cmidrule{1-5}
PointGPT-S & 96.8±2.0 & 98.6±1.1 & 92.6±4.6 & 95.2±3.4 \\
\textbf{+PointACL} & \textbf{97.1±2.3} & \textbf{98.8±1.3} & \textbf{93.0±4.0} & \textbf{95.6±3.0} \\ \bottomrule
\end{tabular}
}
\vspace{-10pt}
\end{table}

\begin{table}[t]
\centering
\caption{\textbf{Part segmentation performance on the ShapeNetPart dataset.} We report the mean Intersection over Union (mIoU) across instances (Ins.) and classes (Cls.).}
\label{table:seg}
\resizebox{\linewidth}{!}{
\begin{tabular}{lcc}
\toprule
Methods & Ins. mIoU & Cls.mIoU \\ \cmidrule{1-3}
\multicolumn{3}{c}{\emph{Supervised Learning   Only}} \\ \cmidrule{1-3}
PointNet~\citep{qi2017pointnet} & 83.7 & 80.4 \\
PointNet++~\citep{qi2017pointnet++} & 85.1 & 81.9 \\
DGCNN~\citep{wang2019dynamic} & 85.2 & 82.3 \\ \cmidrule{1-3}
\multicolumn{3}{c}{\emph{with Self-Supervised   Representation Learning}} \\ \cmidrule{1-3}
MaskPoint~\citep{liu2022masked} & 86.0 & 84.4 \\
GPM~\citep{li2024general} & 85.8 & 84.2 \\
PointMamba~\citep{liang2024pointmamba} & 86.0 & 84.4 \\
\cmidrule{1-3}
Point-BERT~\citep{yu2022point} & 85.6 & 84.1 \\
\textbf{+PointACL} & \hspace{0.90cm} \textbf{86.0} \textcolor{blue}{(+0.4)} & \hspace{0.90cm} \textbf{84.6} \textcolor{blue}{(+0.5)} \\
\cmidrule{1-3}
Point-MAE~\citep{pang2022masked} & 86.1 & 84.2 \\
\textbf{+PointACL} & \hspace{0.90cm} \textbf{86.4} \textcolor{blue}{(+0.3)} & \hspace{0.90cm} \textbf{85.2} \textcolor{blue}{(+1.0)} \\
\cmidrule{1-3}
PointGPT-S~\citep{chen2024pointgpt} & 86.2 & 84.1 \\
\textbf{+PointACL} & \hspace{0.90cm} \textbf{86.7} \textcolor{blue}{(+0.5)} &  \hspace{0.90cm} \textbf{84.8} \textcolor{blue}{(+0.7)} \\
\bottomrule
\end{tabular}
}
\vspace{-10pt}
\end{table}

\noindent\textbf{Few-shot classification on ModelNet40.}
Our PointACL framework was evaluated on the ModelNet40 dataset under few-shot learning settings, and the results are presented in Table~\ref{table:few}. Following standard practice, we carry out 10 separate experiments for each setting and reported mean accuracy along with the standard deviation. Compared to both supervised learning methods and other self-supervised representation learning approaches, PointACL consistently achieves higher accuracy. In the 5-way 10-shot task, our method attains an accuracy of 97.1\% with a standard deviation of 2.3\%, outperforming previous methods. Similarly, in the 10-way 20-shot setting, PointACL achieves an accuracy of 95.6\%, demonstrating superior generalization with limited labeled data. 

\noindent\textbf{Part segmentation on ShapeNetPart.}
We evaluated the effectiveness of our PointACL framework on the part segmentation task using the ShapeNetPart dataset, as shown in Table~\ref{table:seg}. PointACL achieves superior performance compared to both traditional supervised models like PointNet and DGCNN and recent self-supervised methods like Point-MAE and PointGPT-S. Specifically, our method attains an instance mIoU of 86.4\% and a class mIoU of 85.2\% with Point-MAE. These results demonstrate that our attention-driven contrastive learning strategy effectively enhances the model's ability to segment parts in complex 3D shapes, confirming the efficacy of PointACL in advancing the state-of-the-art in point cloud segmentation.

\subsection{Ablation Studies}
\noindent\textbf{Components Analysis.}
We conduct extensive experiments with PointGPT-S on ScanObjectNN to validate the effectiveness of each component. 
Table~\ref{table:Ablation} summarizes the ablation study on different mask strategies and loss functions for the OBJ-BG and OBJ-ONLY settings.
Without masking, the baseline model achieves accuracies of 91.6\% (OBJ-BG) and 90.0\% (OBJ-ONLY). Applying a \emph{Random Mask} slightly improves performance, and adding $\mathcal{L}_{contra}$ further enhances accuracies to 92.1\% and 90.9\%. The \emph{Low-Attention Mask} strategy yields marginal gains, which indicates \emph{Low-Attention Mask} provides the model with more explicit learning guidance compared to a \emph{Random Mask}, encouraging it to leverage key patches to re-evaluate low-attention regions. The \emph{High-Attention Mask} strategy delivers the best results. With $\mathcal{L}_{origin}$ alone, it attains 91.9\% (OBJ-BG) and 91.2\% (OBJ-ONLY). Incorporating $\mathcal{L}_{contra}$ boosts performance to 92.3\% and 91.6\%, the highest in our study. This demonstrates that masking the most informative regions forces the model to learn robust features from less informative areas, and the contrastive loss $\mathcal{L}_{contra}$ enhances feature discrimination.
In summary, the combination of the \emph{High-Attention Mask} strategy and the contrastive loss $\mathcal{L}_{contra}$ significantly improves classification accuracy, highlighting the effectiveness of both components in our method. Please refer to the supplementary material for more ablation studies on hyperparameters.

\begin{table}[t]
\centering
\caption{\textbf{Ablation studies of components in PointACL.} We report the overall accuracy (\%) on ScanObjectNN with PointGPT-S as our backbone. The settings adopted by PointACL are \colorbox{gray!30}{marked}.}
\label{table:Ablation}
\resizebox{\linewidth}{!}{
\begin{tabular}{lcccc}
\toprule
Mask   Strategy & \textit{$\mathcal{L}_{\text{origin}}$} & \textit{$\mathcal{L}_{\text{contra}}$} & OBJ-BG & OBJ-ONLY \\ \cmidrule{1-5}
NO Mask & \checkmark & - & 91.6 & 90.0 \\
Random Mask & \checkmark & - & 91.7 & 90.5 \\
Random Mask & \checkmark & \checkmark & 92.1 & 90.9 \\
Low-Attention Mask & \checkmark & - & 91.7 & 90.7 \\
Low-Attention Mask & \checkmark & \checkmark & 92.0 & 91.4 \\
High-Attention Mask & \checkmark & - & 91.9 & 91.2 \\ \rowcolor[gray]{0.9}
\textbf{High-Attention Mask} & \textbf{\checkmark} & \textbf{\checkmark} & \textbf{92.3} & \textbf{91.6} \\ \bottomrule
\end{tabular}
}
\vspace{-10pt}
\end{table}

\begin{figure}[t]
    \centering
        \includegraphics[width=\linewidth]{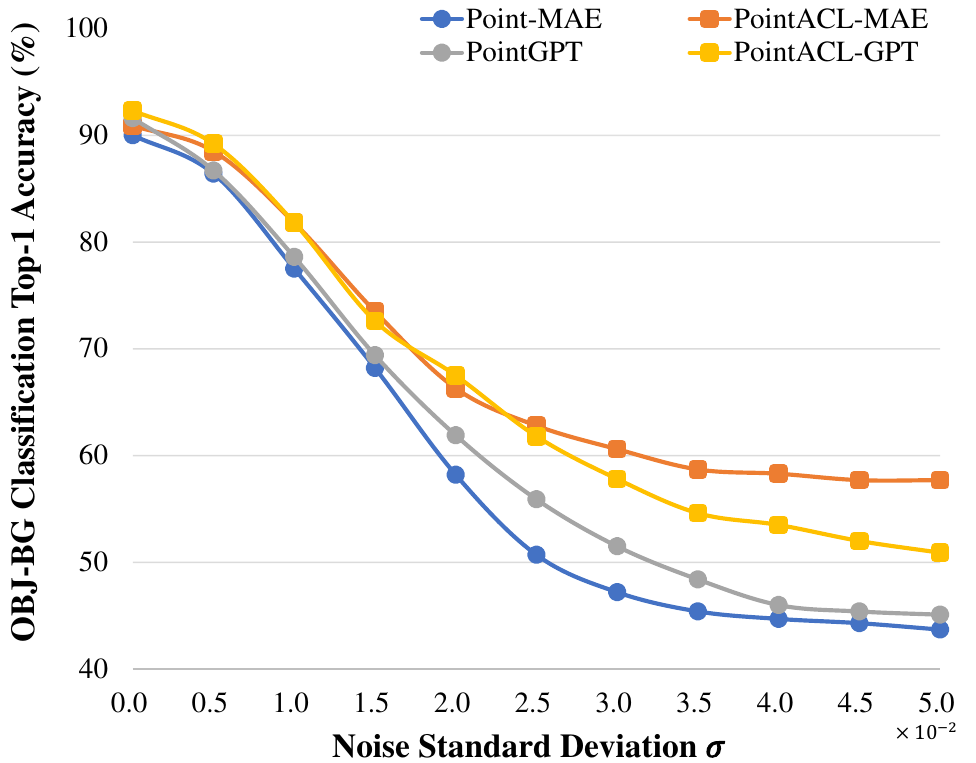}
\caption{\textbf{Gaussian noise analysis on ScanObjectNN.} While the performance of existing methods decmidrules sharply with Gaussian noise, this issue is mitigated by incorporating PointACL.}
    \label{fig: robustness_noise}
    \vspace{-15pt}
\end{figure}

\noindent\textbf{Robustness Analysis.}
To assess the robustness of our PointACL framework, we conducted experiments on the ScanObjectNN dataset under different noisy environments, including Gaussian noise, rotation, scaling, and point dropout, as detailed in Table~\ref{table:Robustness}. Compared to the state-of-the-art models Point-MAE and PointGPT-S, our method consistently achieves higher classification accuracies across both OBJ-BG and OBJ-ONLY settings. For instance, under Gaussian noise with $\sigma=0.03$, PointACL outperforms Point-MAE by up to 13.4\% and PointGPT-S by 6.3\%. Similar improvements are observed with rotational perturbations around the X, Y, and Z axes, scaling factors ranging from 0.5 to 1.5, and point dropout rates of 20\% and 60\%. These results demonstrate that our attention-driven dynamic masking strategy and contrastive learning significantly enhance the model's resilience to noise and transformations. 
Experiments with incremental Gaussian noise ($\sigma$ from 0 to 0.05) on OBJ-BG demonstrate PointACL's superior noise resistance. As illustrated in Figure \ref{fig: robustness_noise}, while all methods' accuracy decreases with increasing noise, PointACL shows notably slower performance degradation, particularly in extreme conditions ($\sigma=0.05$). This enhanced robustness stems from our attention-guided dynamic masking strategy, which promotes the global structure understanding by encouraging focus on under-attended regions.
The performance highlight PointACL's robustness where point clouds often contain noise, occlusions, and varying orientations. 

\begin{table*}[]
\centering
\small
\caption{\textbf{Robustness analysis.} We report the classification accuracy (\%) with four noisy environments: Gaussian noise, rotation, scaling, and droppoint on ScanObjectNN.}
\label{table:Robustness}
\begin{tabular*}{\linewidth}{@{\hspace{0.5em}}@{\extracolsep{\fill}}cccccccccc}
\toprule
\multirow{2}{*}{Dataset} & \multirow{2}{*}{Methods} & \multicolumn{2}{c}{Gaussian Noise} & \multicolumn{3}{c}{Rotation} & Scaling & \multicolumn{2}{c}{DropPoint} \\
\cmidrule{3-10} 
 &  & $\sigma$=0.01 & $\sigma$=0.03 & X[-30 30] & Y[-30 30] & Z[-30 30] & (0.5, 1.5) & 0.2 & 0.6 \\ \cmidrule{1-10}
\multirow{6}{*}{OBJ-BG} & Point-MAE & 77.5 & 47.2 & 72.1 & 87.6 & 72.5 & 86.2 & 87.4 & 84.9 \\
 & \textbf{+PointACL} & 81.8 & 60.6 & 77.3 & 90.5 & 77.3 &  89.3 & 90.7  & 89.7 \\
 & \textcolor{blue}{↑ \textit{Improve}}  & \textcolor{blue}{+4.3} & \textcolor{blue}{+13.4} & \textcolor{blue}{+5.2} & \textcolor{blue}{+2.9} & \textcolor{blue}{+4.8} & \textcolor{blue}{+3.1} & \textcolor{blue}{+3.3} & \textcolor{blue}{+4.8} \\ 
 \cmidrule{2-10} 
 & PointGPT-S & 78.6 & 51.5 & 72.3 & 89.3 & 74.0 & 88.3 & 90.7 & 85.0 \\
 & \textbf{+PointACL} & 81.8 & 57.8 & 76.8 & 91.9 & 79.2 & 90.4 & 91.4 & 86.1 \\
 & \textcolor{blue}{↑ \textit{Improve}}  & \textcolor{blue}{+3.2} & \textcolor{blue}{+6.3} & \textcolor{blue}{+4.5} & \textcolor{blue}{+2.6} & \textcolor{blue}{+5.2} & \textcolor{blue}{+2.1} & \textcolor{blue}{+0.7} & \textcolor{blue}{+1.1} \\ 
 \cmidrule{1-10}
\multirow{6}{*}{OBJ-ONLY} & Point-MAE & 70.9 & 37.0 & 75.4 & 86.7 & 74.9 & 84.0 & 86.6 & 84.5 \\
 & \textbf{+PointACL} & 76.2 & 54.2 & 78.5 & 88.6 & 79.7 & 86.7 & 88.5 & 87.6 \\
 & \textcolor{blue}{↑ \textit{Improve}}  & \textcolor{blue}{+5.3} & \textcolor{blue}{+17.2} & \textcolor{blue}{+3.1} & \textcolor{blue}{+1.9} & \textcolor{blue}{+4.8} & \textcolor{blue}{+2.7} & \textcolor{blue}{+1.9} & \textcolor{blue}{+3.1} \\ 
 \cmidrule{2-10} 
 & PointGPT-S & 71.2 & 39.4 & 72.3 & 89.3 & 74.5 & 86.6 & 89.7 & 85.9 \\
 & \textbf{+PointACL} & 73.3 & 41.3 & 79.9 & 92.3 & 81.8 & 90.0 & 91.2 & 87.4 \\
 & \textcolor{blue}{↑ \textit{Improve}}  & \textcolor{blue}{+2.1} & \textcolor{blue}{+1.9} & \textcolor{blue}{+7.6} & \textcolor{blue}{+3.0} & \textcolor{blue}{+7.3} & \textcolor{blue}{+3.4} & \textcolor{blue}{+1.5} & \textcolor{blue}{+1.5} \\ 
 \bottomrule
\end{tabular*}
\vspace{-10pt}
\end{table*}

\begin{figure}[t]
    \centering
        \includegraphics[width=\linewidth]{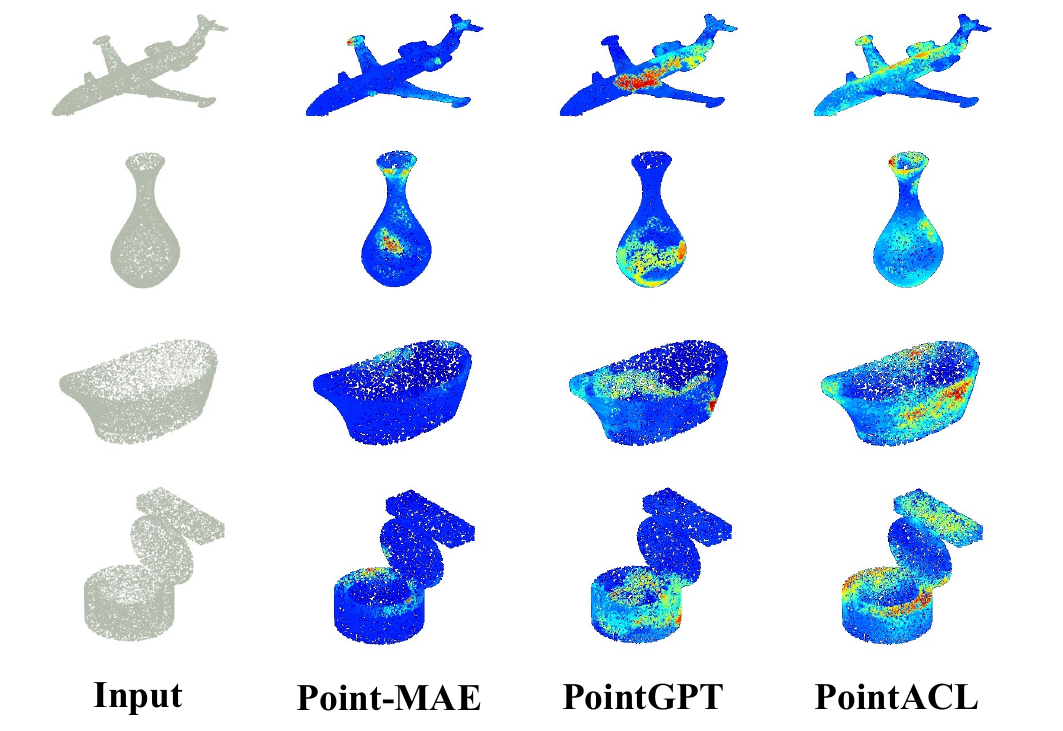}
\caption{\textbf{Attention visualization of PointACL with Point-MAE and PointGPT.} Patches with high attention are closer to red, while patches with low attention are closer to blue. Point-MAE is employed as the backbone of our proposed PointACL.}
    \label{fig: vis}
    \vspace{-12pt}
\end{figure}

\noindent\textbf{Qualitative Analysis.}
As shown in Figure \ref{fig: vis}, the classification heatmaps reveal distinct attention patterns among Point-MAE, PointGPT, and PointACL. While Point-MAE and PointGPT concentrate primarily on high-attention regions, potentially overlooking valuable information elsewhere, PointACL demonstrates more balanced activation across both prominent and subtle areas of the input data. This comprehensive attention distribution directly addresses the limitations identified in existing Transformer-based models, which tend to overlook latent information in less prominent regions. Through our attention-driven dynamic masking strategy and contrastive learning approach, PointACL effectively encourages the model to focus on under-attended regions, thus enhancing its ability to capture global structural information and improve feature discrimination. The richer and more evenly distributed activations in PointACL's heatmaps substantiate its superior capacity for comprehensive point cloud analysis.

\section{Conclusion}
In this work, we present PointACL, an attention-driven contrastive learning framework. By integrating an attention-driven dynamic masking strategy with contrastive learning, our method leverages the model's inherent attention distribution to dynamically mask high-attention regions. This approach guides the network to focus on under-attended low-attention areas, enabling it to learn more comprehensive and robust point cloud feature representations.
Our extensive experiments demonstrate that PointACL significantly enhances the understanding of global structures in point clouds, leading to notable improvements across various tasks, including object classification, part segmentation, and few-shot learning. 
We hope that our work can inspire more explorations of self-supervised learning and contrastive learning in point cloud understanding.
{
    \small
    \bibliographystyle{ieeenat_fullname}
    \bibliography{main}

\begin{thebibliography}{66}
\providecommand{\natexlab}[1]{#1}
\providecommand{\url}[1]{\texttt{#1}}
\expandafter\ifx\csname urlstyle\endcsname\relax
  \providecommand{\doi}[1]{doi: #1}\else
  \providecommand{\doi}{doi: \begingroup \urlstyle{rm}\Url}\fi

\bibitem[Abdelfattah et~al.(2024)Abdelfattah, Hassan, and Alahi]{abdelfattah2024maskclr}
Mohamed Abdelfattah, Mariam Hassan, and Alexandre Alahi.
\newblock Maskclr: Attention-guided contrastive learning for robust action representation learning.
\newblock In \emph{Proceedings of the IEEE/CVF Conference on Computer Vision and Pattern Recognition}, pages 18678--18687, 2024.

\bibitem[Achlioptas et~al.(2018)Achlioptas, Diamanti, Mitliagkas, and Guibas]{achlioptas2018learning}
Panos Achlioptas, Olga Diamanti, Ioannis Mitliagkas, and Leonidas Guibas.
\newblock Learning representations and generative models for 3d point clouds.
\newblock In \emph{International conference on machine learning}, pages 40--49. PMLR, 2018.

\bibitem[Arena et~al.(2022)Arena, Collotta, Pau, and Termine]{arena2022overview}
Fabio Arena, Mario Collotta, Giovanni Pau, and Francesco Termine.
\newblock An overview of augmented reality.
\newblock \emph{Computers}, 11\penalty0 (2):\penalty0 28, 2022.

\bibitem[Bao et~al.(2021)Bao, Dong, Piao, and Wei]{bao2021beit}
Hangbo Bao, Li Dong, Songhao Piao, and Furu Wei.
\newblock Beit: Bert pre-training of image transformers.
\newblock \emph{arXiv preprint arXiv:2106.08254}, 2021.

\bibitem[Chen et~al.(2024)Chen, Wang, Yang, Yu, Yuan, and Yue]{chen2024pointgpt}
Guangyan Chen, Meiling Wang, Yi Yang, Kai Yu, Li Yuan, and Yufeng Yue.
\newblock Pointgpt: Auto-regressively generative pre-training from point clouds.
\newblock \emph{Advances in Neural Information Processing Systems}, 36, 2024.

\bibitem[Chen et~al.(2020)Chen, Liu, Feng, Vallespi-Gonzalez, and Wellington]{chen20203d}
Siheng Chen, Baoan Liu, Chen Feng, Carlos Vallespi-Gonzalez, and Carl Wellington.
\newblock 3d point cloud processing and learning for autonomous driving: Impacting map creation, localization, and perception.
\newblock \emph{IEEE Signal Processing Magazine}, 38\penalty0 (1):\penalty0 68--86, 2020.

\bibitem[Chen et~al.(2017)Chen, Ma, Wan, Li, and Xia]{chen2017multi}
Xiaozhi Chen, Huimin Ma, Ji Wan, Bo Li, and Tian Xia.
\newblock Multi-view 3d object detection network for autonomous driving.
\newblock In \emph{Proceedings of the IEEE conference on Computer Vision and Pattern Recognition}, pages 1907--1915, 2017.

\bibitem[Cheng et~al.(2021)Cheng, Chen, He, Liu, and Bai]{cheng2021net}
Silin Cheng, Xiwu Chen, Xinwei He, Zhe Liu, and Xiang Bai.
\newblock Pra-net: Point relation-aware network for 3d point cloud analysis.
\newblock \emph{IEEE Transactions on Image Processing}, 30:\penalty0 4436--4448, 2021.

\bibitem[Devlin(2018)]{devlin2018bert}
Jacob Devlin.
\newblock Bert: Pre-training of deep bidirectional transformers for language understanding.
\newblock \emph{arXiv preprint arXiv:1810.04805}, 2018.

\bibitem[Dong et~al.(2022)Dong, Qi, Zhang, Zhang, Sun, Ge, Yi, and Ma]{dong2022autoencoders}
Runpei Dong, Zekun Qi, Linfeng Zhang, Junbo Zhang, Jianjian Sun, Zheng Ge, Li Yi, and Kaisheng Ma.
\newblock Autoencoders as cross-modal teachers: Can pretrained 2d image transformers help 3d representation learning?
\newblock \emph{arXiv preprint arXiv:2212.08320}, 2022.

\bibitem[Erhan et~al.(2010)Erhan, Courville, Bengio, and Vincent]{erhan2010does}
Dumitru Erhan, Aaron Courville, Yoshua Bengio, and Pascal Vincent.
\newblock Why does unsupervised pre-training help deep learning?
\newblock In \emph{Proceedings of the thirteenth international conference on artificial intelligence and statistics}, pages 201--208. JMLR Workshop and Conference Proceedings, 2010.

\bibitem[Feng et~al.(2024)Feng, Wang, Quan, and Yang]{feng2024shape2scene}
Tuo Feng, Wenguan Wang, Ruijie Quan, and Yi Yang.
\newblock Shape2scene: 3d scene representation learning through pre-training on shape data.
\newblock \emph{arXiv preprint arXiv:2407.10200}, 2024.

\bibitem[Garrido et~al.(2021)Garrido, Rodrigues, Augusto~Sousa, Jacob, and Castro~Silva]{garrido2021point}
Daniel Garrido, Rui Rodrigues, A Augusto~Sousa, Joao Jacob, and Daniel Castro~Silva.
\newblock Point cloud interaction and manipulation in virtual reality.
\newblock In \emph{2021 5th International Conference on Artificial Intelligence and Virtual Reality (AIVR)}, pages 15--20, 2021.

\bibitem[Goodfellow et~al.(2020)Goodfellow, Pouget-Abadie, Mirza, Xu, Warde-Farley, Ozair, Courville, and Bengio]{goodfellow2020generative}
Ian Goodfellow, Jean Pouget-Abadie, Mehdi Mirza, Bing Xu, David Warde-Farley, Sherjil Ozair, Aaron Courville, and Yoshua Bengio.
\newblock Generative adversarial networks.
\newblock \emph{Communications of the ACM}, 63\penalty0 (11):\penalty0 139--144, 2020.

\bibitem[Hamdi et~al.(2021)Hamdi, Giancola, and Ghanem]{hamdi2021mvtn}
Abdullah Hamdi, Silvio Giancola, and Bernard Ghanem.
\newblock Mvtn: Multi-view transformation network for 3d shape recognition.
\newblock In \emph{Proceedings of the IEEE/CVF International Conference on Computer Vision}, pages 1--11, 2021.

\bibitem[Han et~al.(2024)Han, Tang, Wang, and Li]{han2024mamba3d}
Xu Han, Yuan Tang, Zhaoxuan Wang, and Xianzhi Li.
\newblock Mamba3d: Enhancing local features for 3d point cloud analysis via state space model.
\newblock \emph{arXiv preprint arXiv:2404.14966}, 2024.

\bibitem[He et~al.(2022)He, Chen, Xie, Li, Doll{\'a}r, and Girshick]{he2022masked}
Kaiming He, Xinlei Chen, Saining Xie, Yanghao Li, Piotr Doll{\'a}r, and Ross Girshick.
\newblock Masked autoencoders are scalable vision learners.
\newblock In \emph{Proceedings of the IEEE/CVF conference on computer vision and pattern recognition}, pages 16000--16009, 2022.

\bibitem[Hong et~al.(2023)Hong, Chou, and Liu]{hong2023attention}
Cheng-Yao Hong, Yu-Ying Chou, and Tyng-Luh Liu.
\newblock Attention discriminant sampling for point clouds.
\newblock In \emph{Proceedings of the IEEE/CVF International Conference on Computer Vision}, pages 14429--14440, 2023.

\bibitem[Huang et~al.(2023)Huang, Dong, Yang, Huang, Lau, Ouyang, and Zuo]{huang2023clip2point}
Tianyu Huang, Bowen Dong, Yunhan Yang, Xiaoshui Huang, Rynson~WH Lau, Wanli Ouyang, and Wangmeng Zuo.
\newblock Clip2point: Transfer clip to point cloud classification with image-depth pre-training.
\newblock In \emph{Proceedings of the IEEE/CVF International Conference on Computer Vision}, pages 22157--22167, 2023.

\bibitem[Jing et~al.(2020)Jing, Chen, Zhang, He, and Tian]{jing2020self}
Longlong Jing, Yucheng Chen, Ling Zhang, Mingyi He, and Yingli Tian.
\newblock Self-supervised modal and view invariant feature learning.
\newblock \emph{arXiv preprint arXiv:2005.14169}, 2020.

\bibitem[Li et~al.(2018{\natexlab{a}})Li, Chen, and Lee]{li2018so}
Jiaxin Li, Ben~M Chen, and Gim~Hee Lee.
\newblock So-net: Self-organizing network for point cloud analysis.
\newblock In \emph{Proceedings of the IEEE conference on computer vision and pattern recognition}, pages 9397--9406, 2018{\natexlab{a}}.

\bibitem[Li et~al.(2018{\natexlab{b}})Li, Bu, Sun, Wu, Di, and Chen]{li2018pointcnn}
Yangyan Li, Rui Bu, Mingchao Sun, Wei Wu, Xinhan Di, and Baoquan Chen.
\newblock Pointcnn: Convolution on x-transformed points.
\newblock \emph{Advances in neural information processing systems}, 31, 2018{\natexlab{b}}.

\bibitem[Li et~al.(2024)Li, Gao, Tan, Ren, Yang, and Li]{li2024general}
Zhe Li, Zhangyang Gao, Cheng Tan, Bocheng Ren, Laurence~T Yang, and Stan~Z Li.
\newblock General point model pretraining with autoencoding and autoregressive.
\newblock In \emph{Proceedings of the IEEE/CVF Conference on Computer Vision and Pattern Recognition}, pages 20954--20964, 2024.

\bibitem[Liang et~al.(2024)Liang, Zhou, Xu, Zhu, Zou, Ye, Tan, and Bai]{liang2024pointmamba}
Dingkang Liang, Xin Zhou, Wei Xu, Xingkui Zhu, Zhikang Zou, Xiaoqing Ye, Xiao Tan, and Xiang Bai.
\newblock Pointmamba: A simple state space model for point cloud analysis.
\newblock In \emph{Advances in Neural Information Processing Systems}, 2024.

\bibitem[Liu et~al.(2022)Liu, Cai, and Lee]{liu2022masked}
Haotian Liu, Mu Cai, and Yong~Jae Lee.
\newblock Masked discrimination for self-supervised learning on point clouds.
\newblock In \emph{European Conference on Computer Vision}, pages 657--675. Springer, 2022.

\bibitem[Liu et~al.(2024)Liu, Yu, Wang, Zheng, Deng, Ye, and Wang]{liu2024point}
Jiuming Liu, Ruiji Yu, Yian Wang, Yu Zheng, Tianchen Deng, Weicai Ye, and Hesheng Wang.
\newblock Point mamba: A novel point cloud backbone based on state space model with octree-based ordering strategy.
\newblock \emph{arXiv preprint arXiv:2403.06467}, 2024.

\bibitem[Luppino et~al.(2021)Luppino, Kampffmeyer, Bianchi, Moser, Serpico, Jenssen, and Anfinsen]{luppino2021deep}
Luigi~Tommaso Luppino, Michael Kampffmeyer, Filippo~Maria Bianchi, Gabriele Moser, Sebastiano~Bruno Serpico, Robert Jenssen, and Stian~Normann Anfinsen.
\newblock Deep image translation with an affinity-based change prior for unsupervised multimodal change detection.
\newblock \emph{IEEE Transactions on Geoscience and Remote Sensing}, 60:\penalty0 1--22, 2021.

\bibitem[Ma et~al.(2022)Ma, Qin, You, Ran, and Fu]{ma2022rethinking}
Xu Ma, Can Qin, Haoxuan You, Haoxi Ran, and Yun Fu.
\newblock Rethinking network design and local geometry in point cloud: A simple residual mlp framework.
\newblock \emph{arXiv preprint arXiv:2202.07123}, 2022.

\bibitem[Min et~al.(2022)Min, Zhao, Xiao, Nie, and Dai]{min2022voxel}
Chen Min, Dawei Zhao, Liang Xiao, Yiming Nie, and Bin Dai.
\newblock Voxel-mae: Masked autoencoders for pre-training large-scale point clouds.
\newblock \emph{arXiv preprint arXiv:2206.09900}, 3, 2022.

\bibitem[Misra and Maaten(2020)]{misra2020self}
Ishan Misra and Laurens van~der Maaten.
\newblock Self-supervised learning of pretext-invariant representations.
\newblock In \emph{Proceedings of the IEEE/CVF conference on computer vision and pattern recognition}, pages 6707--6717, 2020.

\bibitem[Navaneet et~al.(2020)Navaneet, Mathew, Kashyap, Hung, Jampani, and Babu]{navaneet2020image}
KL Navaneet, Ansu Mathew, Shashank Kashyap, Wei-Chih Hung, Varun Jampani, and R~Venkatesh Babu.
\newblock From image collections to point clouds with self-supervised shape and pose networks.
\newblock In \emph{Proceedings of the IEEE/CVF Conference on Computer Vision and Pattern Recognition}, pages 1132--1140, 2020.

\bibitem[Pang et~al.(2022)Pang, Wang, Tay, Liu, Tian, and Yuan]{pang2022masked}
Yatian Pang, Wenxiao Wang, Francis~EH Tay, Wei Liu, Yonghong Tian, and Li Yuan.
\newblock Masked autoencoders for point cloud self-supervised learning.
\newblock In \emph{European conference on computer vision}, pages 604--621. Springer, 2022.

\bibitem[Qi et~al.(2017{\natexlab{a}})Qi, Su, Mo, and Guibas]{qi2017pointnet}
Charles~R Qi, Hao Su, Kaichun Mo, and Leonidas~J Guibas.
\newblock Pointnet: Deep learning on point sets for 3d classification and segmentation.
\newblock In \emph{Proceedings of the IEEE conference on computer vision and pattern recognition}, pages 652--660, 2017{\natexlab{a}}.

\bibitem[Qi et~al.(2017{\natexlab{b}})Qi, Yi, Su, and Guibas]{qi2017pointnet++}
Charles~Ruizhongtai Qi, Li Yi, Hao Su, and Leonidas~J Guibas.
\newblock Pointnet++: Deep hierarchical feature learning on point sets in a metric space.
\newblock \emph{Advances in neural information processing systems}, 30, 2017{\natexlab{b}}.

\bibitem[Qi et~al.(2023)Qi, Dong, Fan, Ge, Zhang, Ma, and Yi]{qi2023contrast}
Zekun Qi, Runpei Dong, Guofan Fan, Zheng Ge, Xiangyu Zhang, Kaisheng Ma, and Li Yi.
\newblock Contrast with reconstruct: Contrastive 3d representation learning guided by generative pretraining.
\newblock In \emph{International Conference on Machine Learning}, pages 28223--28243. PMLR, 2023.

\bibitem[Qi et~al.(2025)Qi, Dong, Zhang, Geng, Han, Ge, Yi, and Ma]{qi2025shapellm}
Zekun Qi, Runpei Dong, Shaochen Zhang, Haoran Geng, Chunrui Han, Zheng Ge, Li Yi, and Kaisheng Ma.
\newblock Shapellm: Universal 3d object understanding for embodied interaction.
\newblock In \emph{European Conference on Computer Vision}, pages 214--238. Springer, 2025.

\bibitem[Qian et~al.(2022)Qian, Li, Peng, Mai, Hammoud, Elhoseiny, and Ghanem]{qian2022pointnext}
Guocheng Qian, Yuchen Li, Houwen Peng, Jinjie Mai, Hasan Hammoud, Mohamed Elhoseiny, and Bernard Ghanem.
\newblock Pointnext: Revisiting pointnet++ with improved training and scaling strategies.
\newblock \emph{Advances in neural information processing systems}, 35:\penalty0 23192--23204, 2022.

\bibitem[Qian et~al.(2021)Qian, Meng, Gong, Yang, Wang, Belongie, and Cui]{qian2021spatiotemporal}
Rui Qian, Tianjian Meng, Boqing Gong, Ming-Hsuan Yang, Huisheng Wang, Serge Belongie, and Yin Cui.
\newblock Spatiotemporal contrastive video representation learning.
\newblock In \emph{Proceedings of the IEEE/CVF conference on computer vision and pattern recognition}, pages 6964--6974, 2021.

\bibitem[Radford(2018)]{radford2018improving}
Alec Radford.
\newblock Improving language understanding by generative pre-training.
\newblock 2018.

\bibitem[Ran et~al.(2022)Ran, Liu, and Wang]{ran2022surface}
Haoxi Ran, Jun Liu, and Chengjie Wang.
\newblock Surface representation for point clouds.
\newblock In \emph{Proceedings of the IEEE/CVF conference on computer vision and pattern recognition}, pages 18942--18952, 2022.

\bibitem[Saito and Poovvancheri(2024)]{saito2024point}
Ayumu Saito and Jiju Poovvancheri.
\newblock Point-jepa: A joint embedding predictive architecture for self-supervised learning on point cloud.
\newblock \emph{arXiv preprint arXiv:2404.16432}, 2024.

\bibitem[Sarzynska-Wawer et~al.(2021)Sarzynska-Wawer, Wawer, Pawlak, Szymanowska, Stefaniak, Jarkiewicz, and Okruszek]{sarzynska2021detecting}
Justyna Sarzynska-Wawer, Aleksander Wawer, Aleksandra Pawlak, Julia Szymanowska, Izabela Stefaniak, Michal Jarkiewicz, and Lukasz Okruszek.
\newblock Detecting formal thought disorder by deep contextualized word representations.
\newblock \emph{Psychiatry Research}, 304:\penalty0 114135, 2021.

\bibitem[Sauder and Sievers(2019)]{sauder2019self}
Jonathan Sauder and Bjarne Sievers.
\newblock Self-supervised deep learning on point clouds by reconstructing space.
\newblock \emph{Advances in Neural Information Processing Systems}, 32, 2019.

\bibitem[Tan et~al.(2001)Tan, Robertson, and Czerwinski]{tan2001exploring}
Desney~S Tan, George~G Robertson, and Mary Czerwinski.
\newblock Exploring 3d navigation: combining speed-coupled flying with orbiting.
\newblock In \emph{Proceedings of the SIGCHI conference on Human factors in computing systems}, pages 418--425, 2001.

\bibitem[Uy et~al.(2019)Uy, Pham, Hua, Nguyen, and Yeung]{uy2019revisiting}
Mikaela~Angelina Uy, Quang-Hieu Pham, Binh-Son Hua, Thanh Nguyen, and Sai-Kit Yeung.
\newblock Revisiting point cloud classification: A new benchmark dataset and classification model on real-world data.
\newblock In \emph{Proceedings of the IEEE/CVF international conference on computer vision}, pages 1588--1597, 2019.

\bibitem[Wang et~al.(2021)Wang, Liu, Yue, Lasenby, and Kusner]{wang2021unsupervised}
Hanchen Wang, Qi Liu, Xiangyu Yue, Joan Lasenby, and Matt~J Kusner.
\newblock Unsupervised point cloud pre-training via occlusion completion.
\newblock In \emph{Proceedings of the IEEE/CVF international conference on computer vision}, pages 9782--9792, 2021.

\bibitem[Wang et~al.(2019)Wang, Sun, Liu, Sarma, Bronstein, and Solomon]{wang2019dynamic}
Yue Wang, Yongbin Sun, Ziwei Liu, Sanjay~E Sarma, Michael~M Bronstein, and Justin~M Solomon.
\newblock Dynamic graph cnn for learning on point clouds.
\newblock \emph{ACM Transactions on Graphics (tog)}, 38\penalty0 (5):\penalty0 1--12, 2019.

\bibitem[Wang et~al.(2022)Wang, Yu, Rao, Zhou, and Lu]{wang2022p2p}
Ziyi Wang, Xumin Yu, Yongming Rao, Jie Zhou, and Jiwen Lu.
\newblock P2p: Tuning pre-trained image models for point cloud analysis with point-to-pixel prompting.
\newblock \emph{Advances in neural information processing systems}, 35:\penalty0 14388--14402, 2022.

\bibitem[Wang et~al.(2024)Wang, Chen, Wu, Zhao, Zhou, and Xu]{wang2024pointramba}
Zicheng Wang, Zhenghao Chen, Yiming Wu, Zhen Zhao, Luping Zhou, and Dong Xu.
\newblock Pointramba: A hybrid transformer-mamba framework for point cloud analysis.
\newblock \emph{arXiv preprint arXiv:2405.15463}, 2024.

\bibitem[Wu et~al.(2024)Wu, Jiang, Wang, Liu, Liu, Qiao, Ouyang, He, and Zhao]{wu2024point}
Xiaoyang Wu, Li Jiang, Peng-Shuai Wang, Zhijian Liu, Xihui Liu, Yu Qiao, Wanli Ouyang, Tong He, and Hengshuang Zhao.
\newblock Point transformer v3: Simpler faster stronger.
\newblock In \emph{Proceedings of the IEEE/CVF Conference on Computer Vision and Pattern Recognition}, pages 4840--4851, 2024.

\bibitem[Wu et~al.(2015)Wu, Song, Khosla, Yu, Zhang, Tang, and Xiao]{wu20153d}
Zhirong Wu, Shuran Song, Aditya Khosla, Fisher Yu, Linguang Zhang, Xiaoou Tang, and Jianxiong Xiao.
\newblock 3d shapenets: A deep representation for volumetric shapes.
\newblock In \emph{Proceedings of the IEEE conference on computer vision and pattern recognition}, pages 1912--1920, 2015.

\bibitem[Xie et~al.(2020)Xie, Gu, Guo, Qi, Guibas, and Litany]{xie2020pointcontrast}
Saining Xie, Jiatao Gu, Demi Guo, Charles~R Qi, Leonidas Guibas, and Or Litany.
\newblock Pointcontrast: Unsupervised pre-training for 3d point cloud understanding.
\newblock In \emph{Computer Vision--ECCV 2020: 16th European Conference, Glasgow, UK, August 23--28, 2020, Proceedings, Part III 16}, pages 574--591. Springer, 2020.

\bibitem[Xue et~al.(2023)Xue, Gao, Xing, Mart{\'\i}n-Mart{\'\i}n, Wu, Xiong, Xu, Niebles, and Savarese]{xue2023ulip}
Le Xue, Mingfei Gao, Chen Xing, Roberto Mart{\'\i}n-Mart{\'\i}n, Jiajun Wu, Caiming Xiong, Ran Xu, Juan~Carlos Niebles, and Silvio Savarese.
\newblock Ulip: Learning a unified representation of language, images, and point clouds for 3d understanding.
\newblock In \emph{Proceedings of the IEEE/CVF conference on computer vision and pattern recognition}, pages 1179--1189, 2023.

\bibitem[Xue et~al.(2024)Xue, Yu, Zhang, Panagopoulou, Li, Mart{\'\i}n-Mart{\'\i}n, Wu, Xiong, Xu, Niebles, et~al.]{xue2024ulip}
Le Xue, Ning Yu, Shu Zhang, Artemis Panagopoulou, Junnan Li, Roberto Mart{\'\i}n-Mart{\'\i}n, Jiajun Wu, Caiming Xiong, Ran Xu, Juan~Carlos Niebles, et~al.
\newblock Ulip-2: Towards scalable multimodal pre-training for 3d understanding.
\newblock In \emph{Proceedings of the IEEE/CVF Conference on Computer Vision and Pattern Recognition}, pages 27091--27101, 2024.

\bibitem[Yi et~al.(2016)Yi, Kim, Ceylan, Shen, Yan, Su, Lu, Huang, Sheffer, and Guibas]{yi2016scalable}
Li Yi, Vladimir~G Kim, Duygu Ceylan, I-Chao Shen, Mengyan Yan, Hao Su, Cewu Lu, Qixing Huang, Alla Sheffer, and Leonidas Guibas.
\newblock A scalable active framework for region annotation in 3d shape collections.
\newblock \emph{ACM Transactions on Graphics (ToG)}, 35\penalty0 (6):\penalty0 1--12, 2016.

\bibitem[Yu et~al.(2017)Yu, Zhang, Wang, and Yu]{yu2017seqgan}
Lantao Yu, Weinan Zhang, Jun Wang, and Yong Yu.
\newblock Seqgan: Sequence generative adversarial nets with policy gradient.
\newblock In \emph{Proceedings of the AAAI conference on artificial intelligence}, 2017.

\bibitem[Yu et~al.(2022)Yu, Tang, Rao, Huang, Zhou, and Lu]{yu2022point}
Xumin Yu, Lulu Tang, Yongming Rao, Tiejun Huang, Jie Zhou, and Jiwen Lu.
\newblock Point-bert: Pre-training 3d point cloud transformers with masked point modeling.
\newblock In \emph{Proceedings of the IEEE/CVF conference on computer vision and pattern recognition}, pages 19313--19322, 2022.

\bibitem[Zha et~al.(2024)Zha, Ji, Li, Li, Dai, Chen, Wang, and Xia]{zha2024towards}
Yaohua Zha, Huizhen Ji, Jinmin Li, Rongsheng Li, Tao Dai, Bin Chen, Zhi Wang, and Shu-Tao Xia.
\newblock Towards compact 3d representations via point feature enhancement masked autoencoders.
\newblock In \emph{Proceedings of the AAAI Conference on Artificial Intelligence}, pages 6962--6970, 2024.

\bibitem[Zhang et~al.(2022{\natexlab{a}})Zhang, Guo, Gao, Fang, Zhao, Wang, Qiao, and Li]{zhang2022point}
Renrui Zhang, Ziyu Guo, Peng Gao, Rongyao Fang, Bin Zhao, Dong Wang, Yu Qiao, and Hongsheng Li.
\newblock Point-m2ae: multi-scale masked autoencoders for hierarchical point cloud pre-training.
\newblock \emph{Advances in neural information processing systems}, 35:\penalty0 27061--27074, 2022{\natexlab{a}}.

\bibitem[Zhang et~al.(2022{\natexlab{b}})Zhang, Guo, Zhang, Li, Miao, Cui, Qiao, Gao, and Li]{zhang2022pointclip}
Renrui Zhang, Ziyu Guo, Wei Zhang, Kunchang Li, Xupeng Miao, Bin Cui, Yu Qiao, Peng Gao, and Hongsheng Li.
\newblock Pointclip: Point cloud understanding by clip.
\newblock In \emph{Proceedings of the IEEE/CVF conference on computer vision and pattern recognition}, pages 8552--8562, 2022{\natexlab{b}}.

\bibitem[Zhang et~al.(2023)Zhang, Wang, Qiao, Gao, and Li]{zhang2023learning}
Renrui Zhang, Liuhui Wang, Yu Qiao, Peng Gao, and Hongsheng Li.
\newblock Learning 3d representations from 2d pre-trained models via image-to-point masked autoencoders.
\newblock In \emph{Proceedings of the IEEE/CVF Conference on Computer Vision and Pattern Recognition}, pages 21769--21780, 2023.

\bibitem[Zhang et~al.(2024)Zhang, Li, Yuan, Ji, and Yan]{zhang2024point}
Tao Zhang, Xiangtai Li, Haobo Yuan, Shunping Ji, and Shuicheng Yan.
\newblock Point could mamba: Point cloud learning via state space model.
\newblock \emph{arXiv preprint arXiv:2403.00762}, 2024.

\bibitem[Zhang et~al.(2021)Zhang, Girdhar, Joulin, and Misra]{zhang2021self}
Zaiwei Zhang, Rohit Girdhar, Armand Joulin, and Ishan Misra.
\newblock Self-supervised pretraining of 3d features on any point-cloud.
\newblock In \emph{Proceedings of the IEEE/CVF International Conference on Computer Vision}, pages 10252--10263, 2021.

\bibitem[Zheng et~al.(2024)Zheng, Huang, Mei, Hou, Lyu, Dai, Ouyang, and Gong]{zheng2024point}
Xiao Zheng, Xiaoshui Huang, Guofeng Mei, Yuenan Hou, Zhaoyang Lyu, Bo Dai, Wanli Ouyang, and Yongshun Gong.
\newblock Point cloud pre-training with diffusion models.
\newblock In \emph{Proceedings of the IEEE/CVF Conference on Computer Vision and Pattern Recognition}, pages 22935--22945, 2024.

\bibitem[Zhu et~al.(2023{\natexlab{a}})Zhu, Zhang, He, Guo, Zeng, Qin, Zhang, and Gao]{zhu2023pointclip}
Xiangyang Zhu, Renrui Zhang, Bowei He, Ziyu Guo, Ziyao Zeng, Zipeng Qin, Shanghang Zhang, and Peng Gao.
\newblock Pointclip v2: Prompting clip and gpt for powerful 3d open-world learning.
\newblock In \emph{Proceedings of the IEEE/CVF International Conference on Computer Vision}, pages 2639--2650, 2023{\natexlab{a}}.

\bibitem[Zhu et~al.(2023{\natexlab{b}})Zhu, Ma, Chen, Deng, Huang, and Li]{zhu20233d}
Ziyu Zhu, Xiaojian Ma, Yixin Chen, Zhidong Deng, Siyuan Huang, and Qing Li.
\newblock 3d-vista: Pre-trained transformer for 3d vision and text alignment.
\newblock In \emph{Proceedings of the IEEE/CVF International Conference on Computer Vision}, pages 2911--2921, 2023{\natexlab{b}}.

\end{thebibliography}
}

\section{Appendix}

\subsection{Preliminary}
\textbf{Transformer-based self-supervised learning.} Given a point cloud $X\in \mathbb{R}^{P\times 3}$, we utilize Farthest Point Sampling (FPS) and K-Nearest Neighbors (KNN) algorithms to identify $n$ center points $C$ and their corresponding $k$ nearest neighbors, forming $n$ point patches $P$.  Following the previous methods~\citep{pang2022masked, chen2024pointgpt}, each point patch is normalized to integrate local information. A lightweight token embedding module, implemented via PointNet, subsequently transforms these normalized local patches into trainable point tokens $T$. These point tokens, together with positional embeddings, are input into the transformer blocks to produce latent representations $F$. For different tasks, these latent representations are input into task-specific heads, where they are transformed into specific representations adapted to the task. The learning pipeline based on the Transformer architecture is as follows:

\begin{equation}
F = Transformer(T),
\end{equation}
\begin{equation}
R = Head_{Task-Spec.}(F).
\end{equation}
For Point-MAE, $Head_{Task-Spec.}$ denotes the reconstruction head. For PointGPT, $Head_{Task-Spec.}$ denotes the prediction head.


\subsection{Training Strategy with Cost Analysis}
Given that PointACL determines mask patches based on the attention weights of the backbone network, we suggest two strategies for obtaining these attention weights. The first strategy initializes the network with random attention and applies the attention-driven dynamic masking for adaptive attention refinement during subsequent training. Following standard protocol, the model undergoes pre-training for 300 epochs. This approach does not incur any additional training overhead. The second strategy, by contrast, employs attention learned from the standard branch for initialization, aiming to dynamically adjust the model's dependencies in a targeted manner. This method necessitates 300 epochs of pre-training in the standard branch, followed by another 300 epochs in the dual branch, resulting in a total of 600 epochs. 

Furthermore, we introduce PointACL during the fine-tuning phase of downstream tasks to further evaluate the scalability and effectiveness of our approach. Two strategies are employed here as well: one leverages the pre-trained attention for initialization, while the other requires an additional 300 epochs of training to obtain attention learned from the standard branch for initialization.

Our experimental results, presented in Table~\ref{table:cost}, demonstrate the inherent advantages of our PointACL over existing approaches (such as Point-MAE and PointGPT-S) under the same training time and training phase.
For Point-MAE, with 300 pre-training epochs or 300 fine-tuning epochs, PointACL achieves an accuracy of 90.5\% on the OBJ-BG dataset, surpassing Point-MAE's 90.0\% by a margin of 0.5\%. This improvement persists when both methods are trained for 600 epochs during the fine-tuning phase, with PointACL reaching 90.9\% accuracy compared to Point-MAE's 90.0\%. 
Similarly, when evaluating against PointGPT-S, PointACL continues to exhibit superior performance. With both models trained for 300 fine-tuning epochs on OBJ-ONLY, PointACL attains an accuracy of 90.9\%  compared to PointGPT-S's 90.2\%.  Even when the training epochs are extended to 600, PointACL maintains its advantage, achieving 91.6\% accuracy, outperforming PointGPT-S by 1.4\%. On the OBJ-BG dataset, a similar pattern is observed, where PointACL consistently outperforms PointGPT-S regardless of training duration.

The superior performance of PointACL across various datasets, training epochs, and application phases validates the efficacy of our framework. It demonstrates the performance gains of PointACL are not a consequence of longer training times but are a direct result of designed framework contributions—namely, the attention-driven dynamic masking strategy with contrastive learning. By focusing on under-attended regions and enhancing feature discrimination, PointACL effectively captures both global and local features, leading to enhanced robustness and generalization. 

\begin{table}[t]
\centering
\caption{\textbf{Training cost analysis.} We report the classification accuracy (\%) on ScanObjectNN.}
\label{table:cost}
\resizebox{\linewidth}{!}{
\begin{tabular}{cccccc}
\toprule
\multirow{2}{*}{DataSet} & \multirow{2}{*}{Methods} & \multicolumn{2}{c}{Pre-Training Epoch} & \multicolumn{2}{c}{Finetune Epoch}\\ \cmidrule{3-6}
~ & ~ & 300 & 600 & 300 & 600 \\ \cmidrule{1-6}
\multirow{6}{*}{OBJ-BG} & Point-MAE & 90.0  & 90.2  & 90.0  & 90.0\\ 
~ & \textbf{+PointACL} & 90.5  & 90.5  & 90.5  & 90.9  \\ 
~ & \textcolor{blue}{↑   \textit{Improve}} & \textcolor{blue}{+0.5}  & \textcolor{blue}{+0.3}  & \textcolor{blue}{+0.5}  & \textcolor{blue}{+0.9}  \\ \cmidrule{2-6}
~ & PointGPT-S & 91.6  & 91.7  & 91.6  & 91.9  \\
~ & \textbf{+PointACL} & 91.9  & 91.9  & 92.1  & 92.3  \\
~ & \textcolor{blue}{↑   \textit{Improve}} & \textcolor{blue}{+0.3}  & \textcolor{blue}{+0.2}  & \textcolor{blue}{+0.5}  & \textcolor{blue}{+0.4}  \\ \cmidrule{1-6}
\multirow{6}{*}{OBJ-ONLY} & Point-MAE & 88.3  & 88.5  & 88.3  & 88.3  \\ 
~ & \textbf{+PointACL} & 88.8  & 89.8  & 89.2  & 88.8   \\ 
~ & \textcolor{blue}{↑   \textit{Improve}} & \textcolor{blue}{+0.5}  & \textcolor{blue}{+1.3}  & \textcolor{blue}{+0.9}  & \textcolor{blue}{+0.5}  \\ \cmidrule{2-6}
~ & PointGPT-S & 90.0  & 90.2  & 90.0  & 90.2 \\ 
~ & \textbf{+PointACL} & 90.5  & 91.4  & 90.9  & 91.6  \\ 
~ & \textcolor{blue}{↑   \textit{Improve}} & \textcolor{blue}{+0.5}  & \textcolor{blue}{+1.2}  & \textcolor{blue}{+0.9}  & \textcolor{blue}{+1.4}  \\ \bottomrule
\end{tabular}}
\end{table}

\begin{table}[ht]
\centering
\caption{\textbf{Mask strategy analysis.} We report the classification accuracy (\%) on ScanObjectNN.}
\label{table:mask}
\resizebox{\linewidth}{!}{
\begin{tabular}{ccccc}
\toprule
Mask Probability & Mask Ratio & OBJ-BG & OBJ-ONLY \\ 
\cmidrule{1-4}
\multirow{4}{*}{Fixed Mask} & 0.2 & 91.9  & 90.2  \\ 
~ & 0.4 & 91.9  & 90.9  \\ 
~ & 0.6 & 91.9  & 90.9  \\ 
~ & 0.8 & 91.3  & 90.0  \\ \cmidrule{1-4}
\multirow{4}{*}{Dynamic Mask} & 0.2 & 91.7  & 90.9  \\ 
~ & 0.4 & 91.9  & 91.2  \\ 
~ & 0.6 & \textbf{92.3}  & \textbf{91.6}  \\ 
~ & 0.8 & 91.6  & 90.5 \\ \bottomrule
\end{tabular}
}
\end{table}

\begin{table}[tp]
\centering
\caption{\textbf{Ablation studies of hyperparameters in PointACL.} We report the overall accuracy (\%) on ScanObjectNN. The settings adopted by PointACL are \colorbox{gray!30}{marked}.}
\label{table:Ablation}
\begin{minipage}{0.5\linewidth}
\centering
(a) Probability Temperature. \\[4pt]
\label{table:Ablation3}
\resizebox{\linewidth}{!}{
\begin{tabular}{ccc}
\toprule
$\tau_{pro}$ & OBJ-BG & OBJ-ONLY \\ \cmidrule{1-3}
0.3 & 91.6 & 91.4 \\ \rowcolor[gray]{0.9}
\textbf{0.5} & \textbf{92.3} & \textbf{91.6} \\
0.7 & 92.1 & 91.6 \\
0.9 & 91.9 & 91.4 \\ \bottomrule
\end{tabular}
}
\end{minipage}%
\hfill
\begin{minipage}{0.49\linewidth}
\centering
(b) Contrastive Loss Weight. \\[4pt]
\label{table:Ablation4}
\resizebox{\linewidth}{!}{
\begin{tabular}{ccc}
\toprule
$\lambda$ & OBJ-BG & OBJ-ONLY \\ \cmidrule{1-3}
0.4 & 91.7 & 90.9 \\ \rowcolor[gray]{0.9}
\textbf{0.6} & \textbf{92.3} & \textbf{91.6} \\
0.8 & 92.1 & 90.9 \\
1.0 & 91.7 & 91.0 \\ \bottomrule
\end{tabular}
}
\end{minipage}%
\vspace{-4pt}
\end{table}

\subsection{Ablation Study Analysis}
\textbf{Mask strategy.}
We conduct a thorough exploration into the effects of various mask strategies and mask ratios on the classification performance under the OBJ-BG and OBJ-ONLY settings. Two distinct mask strategies are evaluated: High-Attention Mask with Fixed Mask Probability and High-Attention Masking  Dynamic Mask Probability. The detailed experimental results are presented in Table~\ref{table:mask}. 
When employing the High-Attention Mask with Fixed Mask Probability, we observe that increasing the mask ratio from 0.2 to 0.6 leads to improved performance, with accuracies peaking at 91.9\% on OBJ-BG and 90.9\% on OBJ-ONLY when the mask ratio is 0.6. However, further increasing the mask ratio to 0.8 results in a decmidrule in accuracy. This decline indicates that masking an excessive number of patches impairs the model’s capacity to learn effective representations. Therefore, an optimal mask ratio is crucial, balancing data complexity with sufficient information retention to ensure robust classification.

In contrast, the High-Attention Mask with Dynamic Mask Probability demonstrates notable performance improvements. Specifically, at a mask ratio of 0.6, our model attains the highest accuracies of 92.3\% on OBJ-BG and 91.6\% on OBJ-ONLY, outperforming masking strategies with fixed mask probability. The dynamic adjustment of the mask probability based on attention weights allows the model to more effectively target and mask the most prominent regions, thereby compelling it to learn richer features from less attended areas. This dynamic approach enhances the model's ability to capture global structural information and reduces its reliance on a limited set of salient features.

\textbf{Probability temperature.} We further explore the effects of varying the temperature hyperparameter in the dynamic masking probability. Results in Table~\ref{table:Ablation}(a) indicate that setting $\tau_{pro}$ to 0.5 yields the highest classification accuracies, achieving 92.3\% on OBJ-BG and 91.6\% on OBJ-ONLY. This suggests that this temperature value effectively masks the region of higher attention while maintaining a certain level of dynamic selection, allowing the model to improve global understanding.

\textbf{Contrastive loss weight.} The analysis of contrastive loss weight in Table~\ref{table:Ablation}(b) demonstrates that $\lambda=0.6$ strikes the best balance between the original loss and the contrastive loss. This optimal balance maximizes overall performance and enhances accuracy across both datasets. By fine-tuning the loss weights, PointACL effectively leverages contrastive learning to improve global understanding and generalization capabilities while maintaining task-specific performance.

The experimental results confirm the effectiveness of the proposed attention-driven dynamic masking strategy, which enhances feature representation and classification performance by encouraging the model to learn from under-attended regions. This approach addresses the limitations of prior methods that overly focus on prominent local features, improving robustness and generalization in 3D point cloud analysis.

\begin{figure}[t]
    \centering
        \includegraphics[width=\linewidth]{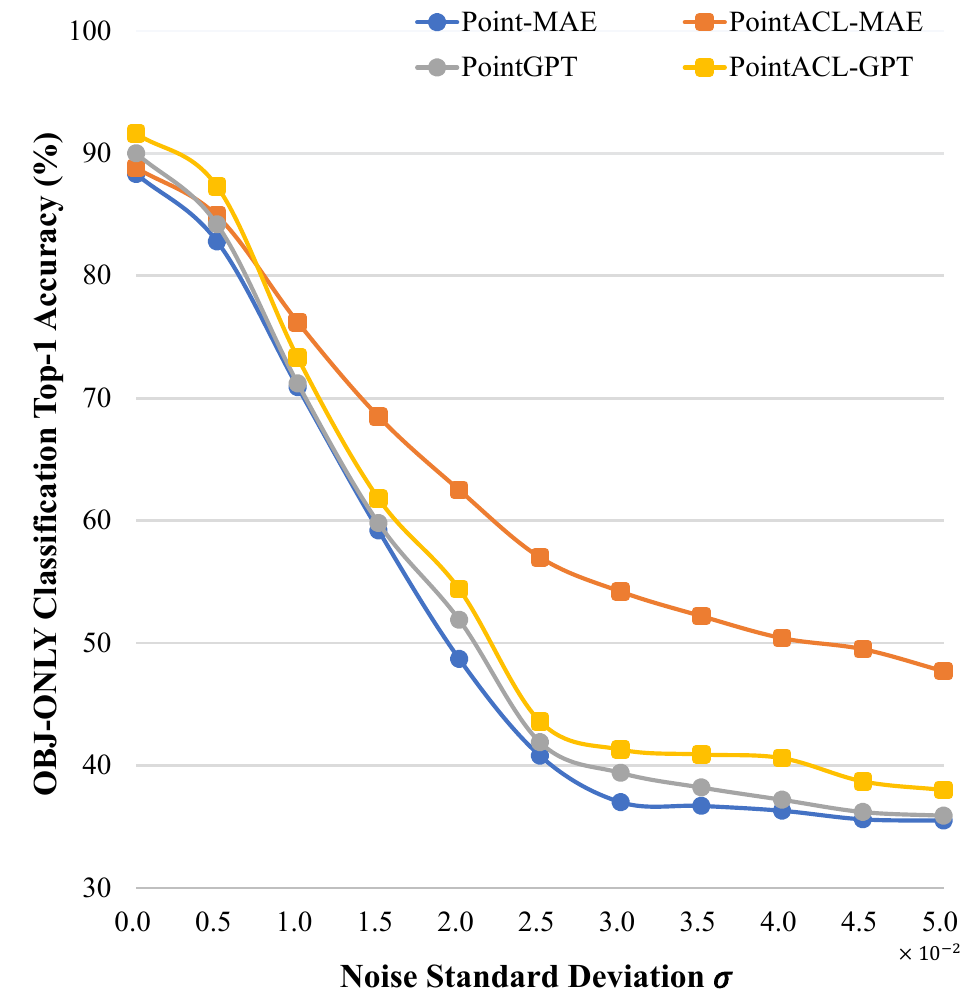}
\caption{\textbf{Gaussian noise analysis on ScanObjectNN.} While the performance of existing methods decmidrules sharply with increasing Gaussian noise, this issue is mitigated by incorporating PointACL. Notably, when Point-MAE is used as the backbone network, our PointACL significantly enhances its robustness, resulting in minimal accuracy degradation.}
    \label{fig: app_noise}
    \vspace{-6pt}
\end{figure}

\begin{figure*}[t]
    \centering
        \includegraphics[width=\linewidth]{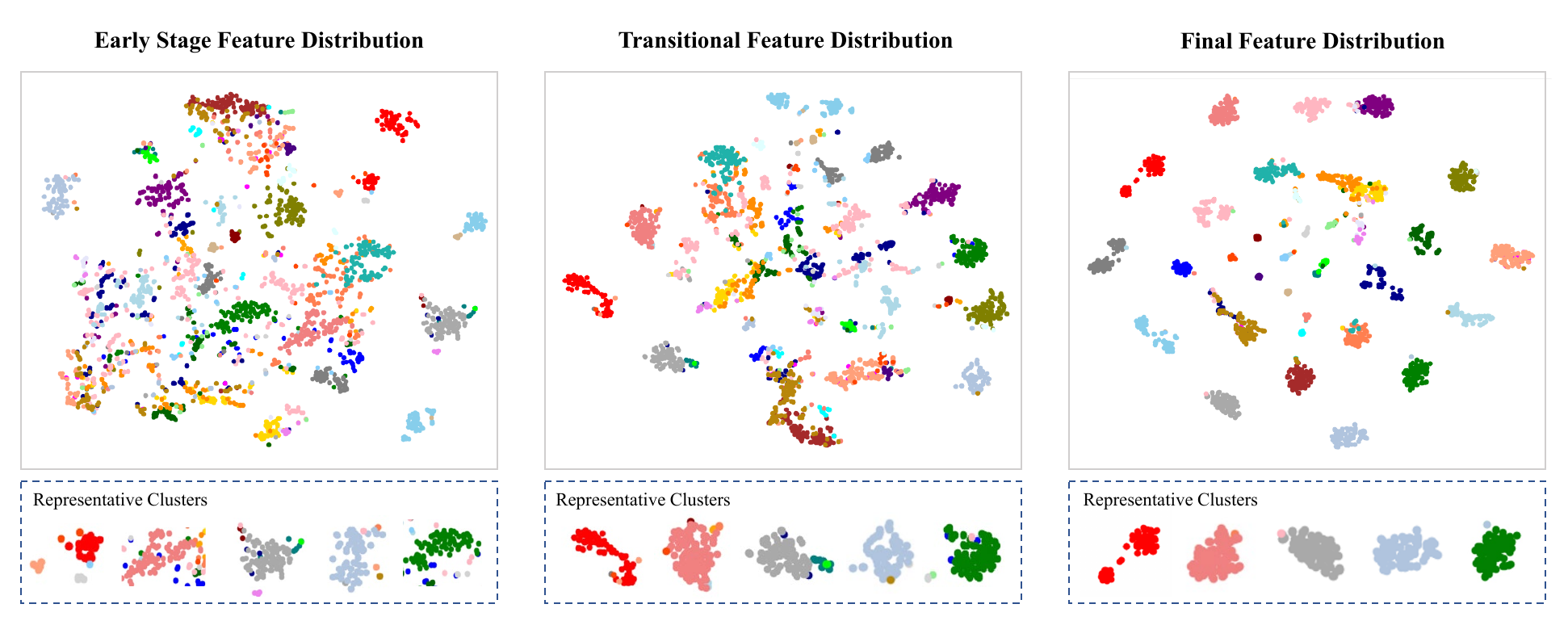}
\caption{\textbf{Feature distribution visualization on ModelNet40.} \textbf{Top:} An overview of the evolution of feature distributions across all 40 classes. \textbf{Bottom:} Detailed depiction of the evolution of feature distributions for selected typical classes.}
    \label{fig: app_tsne}
    \vspace{-6pt}
\end{figure*}

\subsection{Robutness Analysis}

We further evaluate the robustness of our method against existing approaches under Gaussian noise conditions using the OBJ-ONLY subsets of the ScanObjectNN dataset. To simulate noisy point clouds, we add Gaussian noise $X \sim \mathcal{N}(0, \sigma^2)$ to all points, incrementally increasing the noise level by varying $\sigma$ from 0 to 0.05 with step size = 0.005. As illustrated in Figure \ref{fig: app_noise}, while the accuracy of all methods decmidrules as the noise standard deviation $\sigma$ increases, PointACL exhibits a slower performance degradation, demonstrating its superior ability to handle noisy point clouds. Notably, PointACL significantly improves the robustness of the Point-MAE backbone and outperforms baseline methods such as Point-MAE and PointGPT, particularly under extreme noise conditions ($\sigma=0.05$). This improvement can be attributed to our attention-guided dynamic masking strategy, which encourages the model to focus on under-attended regions, thereby enhancing its capacity to capture comprehensive global structural information from point clouds. By not solely relying on salient local features, PointACL mitigates sensitivity to noise-induced perturbations. Additionally, the integration of contrastive learning with the original task further refines feature discrimination, enabling the model to distinguish subtle variations in data even under noisy conditions. The consistently strong performance across both the OBJ-BG and OBJ-ONLY datasets underscores the versatility and reliability of PointACL in diverse settings.

In real-world applications, 3D data is often affected by noise from sensor inaccuracies and environmental factors, making PointACL's robustness to Gaussian noise especially valuable. Its strong performance under such conditions demonstrates its practicality for tasks where data quality is uncertain, underscoring the effectiveness of our framework and its advantage over existing Transformer-based methods.

\subsection{Feature Distribution Evaluation}
Figure \ref{fig: app_tsne} illustrates the evolution of the global feature distribution using t-SNE during the fine-tuning of PointACL, with Point-MAE as the backbone, on the ModelNet40 dataset. In the early stage feature distribution, the feature space is highly scattered with overlapping clusters, indicating that the backbone has not yet learned to effectively discriminate between different classes. As the backbone starts to align global representations from standard branch and masked branch based on attention-driven dynamic masking, the transitional feature distribution shows a notable improvement, with clusters becoming more distinct. However, there still remains some inter-class overlap.

In the final feature distribution, the clusters are well-separated and compact, reflecting a highly discriminative feature space. The backbone has successfully learned to distinguish between different classes with a high degree of accuracy. The representative clusters at the bottom of each visualization further emphasize this progression, showing a clear transition from mixed and overlapping clusters in the early stages to well-defined and isolated clusters in the final stage. These visualizations highlight the effectiveness of the PointACL, demonstrating a clear trajectory of improvement in feature discrimination, culminating in a robust and well-defined feature space.

\subsection{Limatation Analysis}
Despite the significant improvements achieved by PointACL, there are still areas that offer opportunities for further enhancement. For example, while our method has been validated on specific datasets, applying it to a broader range of datasets could further demonstrate its generalizability and robustness. Additionally, although we have shown that PointACL integrates seamlessly with certain Transformer-based architectures, exploring its compatibility with an even wider variety of models could highlight its versatility even more. These considerations open avenues for future research to build upon our work and continue advancing the field of point cloud analysis.

\subsection{Future Works}
While the proposed PointACL framework has shown significant improvements in point cloud analysis tasks, there are several promising directions for future research to further enhance its capabilities and applications. 

One potential avenue is the integration of multi-modal data sources to enrich point cloud representations. By incorporating complementary information from modalities such as images, textual descriptions, or LiDAR intensity values, the model can leverage cross-modal correlations to learn more comprehensive and robust feature embeddings. This multi-modal fusion could enhance the model's ability to understand complex scenes and improve performance in tasks like 3D object detection and semantic segmentation. 
Another direction is the exploration of hierarchical or multi-scale feature learning within the PointACL framework. By capturing features at various spatial resolutions, the model can better represent both local geometric details and global structural contexts. This enhancement could be particularly beneficial for handling large-scale point clouds or scenes with significant variations in point densities.
Additionally, optimizing the computational efficiency of the attention-driven dynamic masking and contrastive learning components is an important consideration for real-world applications. Investigating lightweight architectures or efficient training strategies could make the model more suitable for deployment in resource-constrained environments, such as mobile robots or embedded systems used in autonomous driving.
Lastly, applying the PointACL approach to other types of data representations, such as meshes or voxels, could broaden its applicability across different domains in 3D data processing. Exploring transfer learning techniques between these representations may also provide insights into shared structures and features among various 3D data forms.

By pursuing these future research directions, we aim to further advance the capabilities of PointACL, contributing to the development of more robust, efficient, and versatile models for point cloud analysis. These enhancements have the potential to impact a wide range of applications, including robotics, augmented reality, virtual reality, and autonomous navigation, by enabling more accurate and comprehensive understanding of complex 3D environments.

\end{document}